\documentclass[11pt]{article}

\usepackage{amsmath,amssymb}
\usepackage{amsthm}
\newtheorem{proposition}{Proposition}
\usepackage[margin=1in]{geometry}
\usepackage{graphicx}
\usepackage{booktabs}
\usepackage[hidelinks]{hyperref}
\usepackage{bm}
\usepackage{physics}
\usepackage{xcolor}
\usepackage{tikz}
\usepackage{float, subcaption}
\usepackage{placeins}
\usepackage{microtype}
\numberwithin{equation}{section}

\newcommand{\He}{\operatorname{He}}
\newcommand{\md}{\,\mathrm{d}}

\title{Hyperbolic machine learning moment closures for the Vlasov--Poisson--BGK equations}
\author{{Andrew J. Christlieb\textsuperscript{1},
Sining Gong\textsuperscript{1},
Juntao Huang\textsuperscript{2},}\\
{Nicholas A. Krupansky\textsuperscript{3},
Shiping Zhou\textsuperscript{1}}\\[0.75em]
{\small\textsuperscript{1}Department of Computational Mathematics, Science and Engineering,}\\[-0.1em]
{\small Michigan State University, East Lansing, MI 48824, USA}\\[0.3em]
{\small\textsuperscript{2}Department of Mathematical Sciences,
University of Delaware, Newark, DE 19716, USA}\\[0.3em]
{\small\textsuperscript{3}U.S. Army DEVCOM Ground Vehicle Systems Center, Warren, MI 48397, USA}
}
\date{}

\begin{document}

\maketitle

\begin{abstract}
We develop a hyperbolic machine learning moment closure for the Vlasov--Poisson--Bhatnagar--Gross--Krook (VPBGK) system by extending the characteristic-root gradient closure in Christlieb et al.\ \cite{Christlieb2025a} to self-consistent electrostatic coupling.
For local, field-independent closure coefficients, we show that the periodic Poisson coupling leaves the principal transport matrix unchanged at a given admissible moment state.
The neural network predicts real characteristic roots, which are mapped algebraically to coefficients that express the gradient of the unresolved moment in terms of retained-moment gradients.
A prescribed positive minimum root gap enforces strict hyperbolicity of the transport part.
Numerical tests show that a suitable minimum root gap reduces the number of failure simulations.
Compared with classical moment closures at the same moment order, the model gives smaller errors over most of the tested collision regimes, with clear improvements in higher-order moments.
With network parameters fixed after training on short-time periodic kinetic trajectories, we test benchmark configurations absent from the training data, including Landau damping and a nonperiodic two-beam slab with spatially varying collisionality.
These tests extend far beyond the training time window and use the closure without retraining.
The model retains predictive accuracy at low to moderate Knudsen numbers.
Although the closure construction remains hyperbolic at admissible states, the results for Knudsen number $\mathrm{\tau}>0.1$ show that further improvements in accuracy and numerical robustness are needed.
\end{abstract}

\section{Introduction}
Plasma applications, including Z-pinch configurations, motivates efficient kinetic descriptions across a wide range of spatial scales and collisional regimes \cite{shumlak2020z,hoffmann2023gyrokinetic}.
In this work, we consider the Vlasov--Poisson--Bhatnagar--Gross--Krook (VPBGK) system as a tractable electrostatic setting that couples particle transport, self-consistent electric fields, and collisional relaxation \cite{Bhatnagar1954,Cai2013Vlasov}.
Direct simulation requires resolving the distribution function in both physical and velocity space, which becomes computationally demanding as the dimension and required resolution increase.
Moment models provide a reduced description by evolving a finite set of velocity moments in physical space.
The evolution equations for these retained moments generally depend on unresolved higher-order moments, so a closure is required to obtain a finite system.

Classical moment closure approaches include Grad's moment closure \cite{Grad1949}, regularized moment equations \cite{Struchtrup2003}, globally hyperbolic moment equations (HME) \cite{Cai2012a,Cai2013Vlasov}, filtered hyperbolic moment methods \cite{Di2019}, and quadrature-based moment equations (QBME) \cite{Koellermeier2014,KoellermeierCastro2021}.
At low moment order, these methods may have limited accuracy in nonequilibrium regimes, while increasing the number of retained moments increases the computational cost.
This tradeoff motivates learning moment models from kinetic data to improve accuracy with a fixed number of moments.
Han et al.\ learn generalized moments and the associated dynamics to construct reduced models across a wide range of Knudsen numbers \cite{Han2019}.
Li, Dong, and Wang learn moment closures for the Boltzmann--BGK equation while incorporating Galilean, reflection, and scaling invariances \cite{Li2023}.
Accurate fitting of kinetic data alone, however, does not ensure that the closed evolution equations retain the structural properties of the underlying kinetic model.
Entropy-based closures provide hyperbolic moment systems through a convex entropy structure \cite{Levermore1996}.
Neural approximations seek to preserve this structure through suitable convexity constraints \cite{Schotthofer2022,Porteous2023}.
For unbounded velocity domains, however, a nonnegative distribution may have the prescribed moments even when no distribution with those moments minimizes the entropy \cite{hauck2008convex}.
Direct constraints on the transport spectrum offer another route to hyperbolic learned closures without requiring an entropy-minimizing reconstruction.

Gradient-based learned closures provide such a framework for imposing hyperbolicity constraints on the moment system.
For the linear radiative transfer equation (RTE), Huang et al.\ approximate the spatial gradient of the unresolved moment by a linear combination of retained-moment gradients, with coefficients supplied by a neural network \cite{Huang2022a}.
Their subsequent work derives constraints that make the learned system globally symmetrizable hyperbolic \cite{Huang2023b}.
Huang et al.\ subsequently parameterize the characteristic speeds directly and introduce separate architectures for enforcing bounded speeds or strict hyperbolicity \cite{Huang2023c}.
Christlieb et al.\ extend this approach to nonlinear BGK dynamics by learning characteristic roots and mapping them to gradient-closure coefficients through a deterministic polynomial transformation \cite{Christlieb2025a}.
Distinct real roots give strict transport hyperbolicity at states with positive density and temperature.
More recently, this framework is generalized to the RTE in two dimensions in physical space and two dimensions in angular space \cite{huang2026machine4} and the linear Boltzmann equation with uncertainties \cite{huang2026uncertainty}.
Our present work builds on the characteristic-root construction \cite{Christlieb2025a} to investigate learned moment closures for self-consistent VPBGK system.

Extending this framework to VPBGK requires assessing the closure within self-consistent electric filed dynamics.
Errors in the evolving charge density alter the field through Poisson's equation, and the resulting field feeds back into subsequent moment evolution.
This coupling motivates evaluating closure accuracy and robustness through both moment predictions and electrostatic observables.
Classical hyperbolic moment methods have been applied to Vlasov--Poisson and VPBGK dynamics \cite{Cai2013Vlasov}, and machine-learning approaches to plasma moment closure are reviewed by Burles and Camporeale \cite{burles2026machine}.
Bois et al.\ learn an Euler--Poisson heat-flux closure using spatial profiles of density, velocity, and temperature as neural-network inputs \cite{Bois2022}.
Huang et al.\ use a Fourier neural operator to learn a heat-flux closure for nonlinear Landau damping \cite{Huang2025Landau}.
Barbour et al.\ use reservoir computing to close a Fourier--Hermite representation of Vlasov--Poisson dynamics \cite{Barbour2025}.
Here, we investigate whether a characteristic-root gradient closure whose coefficients depend on instantaneous local retained moments can improve accurate and robust predictions in coupled VPBGK evolution.

We address this question through three contributions.
First, we extend the characteristic-root closure of \cite{Christlieb2025a} to the VPBGK moment equations.
For closure coefficients depending only on the retained moments and a compatible periodic Poisson problem, we show that eliminating the electric field leaves the principal transport matrix unchanged at a fixed admissible moment state.
Consequently, the coupled system retains strict transport hyperbolicity when density and temperature are positive and the characteristic roots are real and distinct.
Second, we enforce a prescribed minimum root separation and assess its effects on accuracy and numerical robustness.
Huang et al.\ enforce root separation through cumulative positive increments \cite{Huang2023c}.
Our implementation sorts the predicted roots and adds centered offsets that preserve their mean.
We study the effects of root separation, moment order, and the AdamW and SOAP optimizers \cite{Loshchilov2019AdamW, Vyas2025SOAP} on closure-fitting errors, solution accuracy, and simulation completion.
Third, we evaluate the selected closure in coupled VPBGK simulations, keeping the trained network parameters fixed throughout.
The closure is trained on short-time periodic kinetic trajectories, with constant collisionality within each trajectory, and uses local retained moments as inputs.
We compare it with Grad \cite{Grad1949}, HME \cite{Cai2012a,Cai2013Vlasov}, and QBME \cite{Koellermeier2014}, and assess moment errors and Landau-damping field histories.
Without retraining, we test the learned closure on benchmark configurations absent from the training data, including Landau damping and a nonperiodic electrostatic slab with two-beam initial data and spatially varying collisionality.
The temporal-extrapolation, slab, and Landau-damping tests all extend far beyond the short time interval used for training.
This simplified slab benchmark is motivated by Z-pinch applications.
Together, these tests demonstrate accuracy improvements over classical closures in the comparison benchmarks and evaluate the fixed closure well beyond its training times and configurations.
Predictive accuracy is retained at low to moderate Knudsen numbers, while further improvements are needed for weak collisions (Knudsen number larger than $0.1$).

The rest of the manuscript is organized as follows.
Section~\ref{sec2:model} introduces the VPBGK model, its Hermite moment hierarchy, and the characteristic-root closure with Poisson coupling.
Section~\ref{sec3:training-and-numerics} describes kinetic data generation, neural-network training and selection, and numerical discretization.
Section~\ref{sec:numerical-tests} presents the classical-closure comparisons, Landau-damping diagnostics, temporal extrapolation, and slab tests.
Section~\ref{sec:conclusions} concludes the paper.

\section{Hyperbolic Moment Closure}
\label{sec2:model}

This section introduces the VPBGK model and its Hermite moment expansion.
For the gradient closures considered here, we show that self-consistent Poisson coupling leaves the principal transport matrix unchanged at fixed moment state.
This allows us to apply the characteristic root construction of \cite{Christlieb2025a} to obtain a hyperbolic gradient closure.

We consider a single-species electrostatic model in one spatial and one velocity dimension, with a periodic spatial domain $\Omega=(0,L)$ and velocity domain $\mathbb R$.
The nonnegative distribution function $f(x,v,t)$ describes the particle density in phase space, which satisfies the nondimensional VPBGK equation \cite{Bhatnagar1954,Cai2013Vlasov},
\begin{equation}
    \partial_t f+v\partial_x f+E\partial_v f
    =\frac{1}{\tau(x)}\bigl(\mathcal M[f]-f\bigr),
    \qquad (x,v)\in\Omega\times\mathbb R.
    \label{eq:vpbgk-kinetic}
\end{equation}
The Vlasov terms describe free streaming and acceleration by the self-consistent electric field $E(x,t)$.
Poisson's equation determines the electric field relative to a constant neutralizing background $\rho_0$,
\begin{equation}
    E=-\partial_x\phi,
    \quad
    -\partial_{xx}\phi=\rho-\rho_0, \quad \rho_0=\frac{1}{L}\int_0^L\rho(x,0)\,\mathrm dx, \quad \rho=\int_{\mathbb R}f\,\mathrm dv
    \label{eq:vpbgk-poisson}
\end{equation}
with periodic $f$ and the electrostatic potential $\phi$.
Mass conservation ensures the compatibility condition $\int_0^L(\rho-\rho_0)\,\mathrm dx=0$ for all time.
Periodicity of $\phi$ gives a mean-zero electric field, and we fix the additive constant in $\phi$ by requiring $\int_0^L\phi(x,t)\,\mathrm dx=0$.
The BGK term describes relaxation toward the local Maxwellian over a prescribed time scale $\tau(x)>0$.
The relaxation time may be constant or spatially varying.
The Maxwellian is determined by the local density $\rho$, mean velocity $u$, and scaled temperature $\theta$,
\begin{equation*}
    \mathcal M[f]
    =\frac{\rho}{\sqrt{2\pi\theta}}
      \exp\left[-\frac{(v-u)^2}{2\theta}\right], \quad u=\frac{1}{\rho}\int_{\mathbb R}vf\,\mathrm dv,
    \quad
    \theta=\frac{1}{\rho}\int_{\mathbb R}(v-u)^2f\,\mathrm dv.
\end{equation*}
We assume $\rho>0$, $\theta>0$, and sufficient velocity decay for the moment calculations below.

Since $\mathcal{M}[f]$ and $f$ have identical first three velocity moments, the BGK term contribution no mass, momentum, or energy source \cite{Bhatnagar1954,Cai2013Vlasov}.
The BGK collision term therefore preserves local mass, momentum, and kinetic energy \cite{Bhatnagar1954,Cai2013Vlasov}.
Multiplying \eqref{eq:vpbgk-kinetic} by $(1,v,\frac12v^2)^{\mathsf T}$ and integrating over velocity gives
\begin{equation}
    \partial_t
    \begin{bmatrix}
        \rho \\
        \rho u \\
        \frac12\rho(u^2+\theta)
    \end{bmatrix}
    +\partial_x
    \begin{bmatrix}
        \rho u \\
        \rho u^2+\rho\theta \\
        \frac12\rho u(u^2+3\theta)+q
    \end{bmatrix}
    =
    \begin{bmatrix}
        0 \\
        \rho E \\
        \rho uE
    \end{bmatrix},
    \label{eq:macroscopic-balances}
\end{equation}
where $q=\frac12\int_{\mathbb R}(v-u)^3f\,\mathrm dv$ is the heat flux.
These balance laws are not closed because $q$ is not determined by $(\rho, u,\theta)$.

\subsection{Hermite expansion and truncated moment system}

Following Grad's Hermite expansion \cite{Grad1949} and its application to VPBGK \cite{Cai2013Vlasov}, we expand the distribution in a locally centered and scaled basis.
Then we show that, under the local gradient closures considered here, Poisson coupling leaves the principal transport matrix unchanged at a fixed admissible state.
With $\xi=(v-u)/\sqrt{\theta}$, the basis is
\begin{equation}
    \mathcal H_k(v,u,\theta)
    =\frac{1}{\sqrt{2\pi}}\theta^{-(k+1)/2}
      \He_k(\xi)e^{-\xi^2/2},
    \label{eq:moving-hermite-basis}
\end{equation}
where $\He_k$ is the probabilists' Hermite polynomial of degree $k$.
The full expansion and its order-$M$ approximation are
\begin{equation}
    f=\sum_{k=0}^{\infty}f_k\mathcal H_k,
    \qquad
    f^{(M)}=\sum_{k=0}^{M}f_k\mathcal H_k.
    \label{eq:hermite-expansion}
\end{equation}

The moment definitions and Hermite orthogonality give
\begin{equation}
    f_0=\rho,
    \qquad
    f_1=f_2=0,
    \qquad
    q=3f_3.
    \label{eq:heat-flux-f3}
\end{equation}
The coefficients $f_k$ for $k\geq3$ describe deviations from the local Maxwellian.
For $M\geq3$, the retained primitive state is
\begin{equation}
    \bm w=(\rho,u,\theta,f_3,\ldots,f_M)^{\mathsf T}
    \in\mathbb R^{M+1},
    \qquad
    \rho>0,\quad\theta>0.
    \label{eq:primitive-hermite-state}
\end{equation}
This gives an $(M+1)$-moment model.

Substituting $q=3f_3$ into \eqref{eq:macroscopic-balances} and expressing the balances in primitive variables gives
\begin{equation}
    \begin{aligned}
        \partial_t\rho+u\partial_x\rho+\rho\partial_xu&=0,\\
        \partial_tu+u\partial_xu
        +\frac{\theta}{\rho}\partial_x\rho+\partial_x\theta&=E,\\
        \partial_t\theta+u\partial_x\theta
        +2\theta\partial_xu+\frac{6}{\rho}\partial_xf_3&=0.
    \end{aligned}
    \label{eq:primitive-density}
\end{equation}
Hermite projection gives the remaining equations for $3\leq k\leq M$ \cite{Cai2013Vlasov},
\begin{equation}
    \begin{aligned}
        \partial_t f_k
        &-\frac{3f_{k-2}}{\rho}\partial_xf_3
        -\frac{\theta f_{k-1}}{\rho}\partial_x\rho
        +\left(
            \frac{k-1}{2}f_{k-1}
            +\frac{\theta}{2}f_{k-3}
        \right)\partial_x\theta\\
        &+(k+1)f_k\partial_xu
        +u\partial_xf_k+\theta\partial_xf_{k-1}
        +(k+1)\partial_xf_{k+1}
        =-\frac{1}{\tau(x)}f_k.
    \end{aligned}
    \label{eq:hermite-hierarchy}
\end{equation}

The moment system requires a closure because its last equation contains the unresolved derivative $\partial_x f_{M+1}$.
We consider local gradient closures that express this derivative as a linear combination of the retained state gradients, with coefficients depending only on $\bm w$, as follows
\begin{equation}
    \partial_x f_{M+1}
    =\sum_{i=0}^{M}\mathcal N_i(\bm w)\partial_xw_i.
    \label{eq:learned-gradient-closure}
\end{equation}
Here $\mathcal{N}_i(\bm w)$ is the scalar closure coefficient.

In the locally centered Hermite basis, substitution of the mean velocity equations cancels the electric acceleration terms from the higher-moment equations.
Consequently, the electric field enters directly only as a source in the velocity equation, while the spatial derivative terms coincide with those of the force-free BGK system, given as following
\begin{equation}
    \begin{aligned}
        \partial_t\bm w+\bm A(\bm w)\partial_x\bm w
        &=\bm Q_{\mathrm{BGK}}(\bm w,\tau)+\bm Q_E(E),\\
        \partial_xE&=\rho-\rho_0,
    \end{aligned}
    \label{eq:closed-vpbgk-system}
\end{equation}
where
\begin{equation}
    \bm Q_{\mathrm{BGK}}
    =(0,0,0,-f_3/\tau,\ldots,-f_M/\tau)^{\mathsf T},
    \qquad
    \bm Q_E=(0,E,0,\ldots,0)^{\mathsf T}.
    \label{eq:vpbgk-sources}
\end{equation}
The following proposition explains why the BGK transport-matrix construction also applies to this coupled system.

\begin{proposition}[Preservation of transport hyperbolicity]
\label{prop:poisson-hyperbolicity}
Consider the closed Hermite moment system \eqref{eq:closed-vpbgk-system}.
Assume that its gradient-closure coefficients depend only on $\bm w$.
At each admissible state $\bm w$, its transport matrix coincides with that of the force-free BGK moment equations derived from the same closure.
Eliminating $E$ through the compatible periodic Poisson problem leaves this first-order principal transport matrix unchanged.
Consequently, the two systems have the same characteristic polynomial and diagonalizability at that state.
In particular, The VPBGK system is strictly hyperbolic wherever the corresponding BGK system is strictly hyperbolic.
\end{proposition}

\noindent\textit{Proof.}
The periodic Poisson solution is $E=\partial_x^{-1}(\rho-\rho_0)$.
Here $\partial_x^{-1}$ is the inverse derivative on zero-mean periodic functions.
Its Fourier multiplier is $(\mathrm i k)^{-1}$ for nonzero wave numbers $k=2\pi n/L$, and its mean mode is zero.
The field map therefore has order $-1$ and enters only through $\bm Q_E$.
The transport operator has order $1$.
Eliminating $E$ leaves the first-order principal symbol unchanged, together with its characteristic polynomial and diagonalizability.
\hfill$\square$

Thus, at a fixed admissible state, the algebraic relations between the gradient closure coefficients and the characteristic polynomial are identical to those in the BGK construction of \cite{Christlieb2025a}.
Section~\ref{subsec:root-to-coefficient} gives a brief review of these relations.
The electric field can change the evolving state and its dispersion and growth rate, but it does not enter this algebraic transformation.

\subsection{Hyperbolic neural gradient closure}
\label{subsec:root-to-coefficient}

We review the algebraic construction of \cite{Christlieb2025a}, which transforms gradient-closure coefficients to characteristic roots.
The following transformations are evaluated at a fixed admissible state $\bm w$.

For the local closure \eqref{eq:learned-gradient-closure}, the classical Grad's closure sets $f_{M+1}=0$, and hence $\partial_{x} f_{M+1}=0$, in \eqref{eq:hermite-hierarchy}.
The resulting transport matrix $\bm A_{\mathrm{Grad}}(\bm w)$ follows from \eqref{eq:primitive-density}--\eqref{eq:hermite-hierarchy} and has the structure
{\small
\renewcommand{\arraystretch}{1.4}
\setlength{\arraycolsep}{2.5pt}
\begin{equation*}
\bm A_{\mathrm{Grad}} =
\begin{pmatrix}
u & \rho & 0 & 0 & 0 & 0 & 0
& \multicolumn{2}{c}{\cdots} & 0 \\
\frac{\theta}{\rho} & u & 1 & 0 & 0 & 0 & 0
& \multicolumn{2}{c}{\cdots} & 0 \\
0 & 2\theta & u & \frac{6}{\rho} & 0 & 0 & 0
& \multicolumn{2}{c}{\cdots} & 0 \\
0 & 4f_3 & \frac{1}{2}\rho\theta & u & 4 & 0 & 0
& \multicolumn{2}{c}{\cdots} & 0 \\
-\frac{\theta}{\rho}f_3 & 5f_4 & \frac{3}{2}f_3
& \theta & u & 5 & 0
& \multicolumn{2}{c}{\cdots} & 0 \\
-\frac{\theta}{\rho}f_4 & 6f_5 & 2f_4
& -\frac{3}{\rho}f_3 & \theta & u & 6
& \multicolumn{2}{c}{\cdots} & 0 \\
-\frac{\theta}{\rho}f_5 & 7f_6
& \frac{\theta}{2}f_3+\frac{5}{2}f_5
& -\frac{3}{\rho}f_4 & 0 & \theta & u
& \multicolumn{2}{c}{\cdots} & 0 \\
\vdots & \vdots & \vdots & \vdots & \vdots
& & \ddots & \ddots & \ddots & 0 \\
-\frac{\theta}{\rho}f_{M-2} & Mf_{M-1}
& \frac{\theta}{2}f_{M-4}+\frac{M-2}{2}f_{M-2}
& -\frac{3}{\rho}f_{M-3}
& 0 & \cdots & 0 & \theta & u & M \\
-\frac{\theta}{\rho}f_{M-1} & (M+1)f_M
& \frac{\theta}{2}f_{M-3}+\frac{M-1}{2}f_{M-1}
& -\frac{3}{\rho}f_{M-2}
& 0 & \multicolumn{2}{c}{\cdots} & 0 & \theta & u
\end{pmatrix}.
\end{equation*}
}
The learned transport matrix $\bm A_{\mathrm{ML}}$, which is the matrix $\bm A$ in \eqref{eq:closed-vpbgk-system}, differs from $\bm A_{\mathrm{Grad}}$ only in its last row:
\begin{equation*}
    \bm A_{\mathrm{ML}}(\bm w)
    =\bm A_{\mathrm{Grad}}(\bm w)
    +(M+1)\bm e_{M+1}\bm{\mathcal N}(\bm w)^{\mathsf T},
    \qquad
    \bm{\mathcal N}=(\mathcal N_0,\ldots,\mathcal N_M)^{\mathsf T},
\end{equation*}
where $\bm e_{M+1}$ is the last standard coordinate vector in $\mathbb R^{M+1}$.

Let $\bm a=(a_0,\ldots,a_M)$ be the last row of $\bm A_{\mathrm{ML}}$.
Its entries satisfy
\begin{equation}
    a_i=(a_{\mathrm{Grad}})_i+(M+1)\mathcal N_i,
    \qquad 0\leq i\leq M,
    \label{eq:closure-to-last-row}
\end{equation}
where $\bm a_{\mathrm{Grad}}$ is the last row of $\bm A_{\mathrm{Grad}}$.
This gives an affine, invertible relation between $\bm{\mathcal N}=(\mathcal N_0,\ldots,\mathcal N_M)^{\mathsf T}$ and $\bm a$.

Writing $y=\lambda-u$, we represent the shifted characteristic polynomial in the temperature-scaled Hermite and monomial bases as
\begin{equation}
    \begin{aligned}
        P_{\bm a}(y)
        &=\det\bigl((u+y)I-\bm A_{\mathrm{ML}}\bigr)\\
        &=\sum_{k=0}^{M+1}\beta_k\theta^{k/2}
          \He_k\!\left(\frac{y}{\sqrt{\theta}}\right)
          =\sum_{k=0}^{M+1}c_k y^k.
    \end{aligned}
    \label{eq:shifted-characteristic-polynomial}
\end{equation}
Here $\beta_k$ and $c_k$ are the Hermite and monomial coefficients, respectively, with $\beta_{M+1}=c_{M+1}=1$.
We write $\bm\beta=(\beta_0,\ldots,\beta_M)^{\mathsf T}$ and $\bm c=(c_0,\ldots,c_M)^{\mathsf T}$ for the remaining coefficients.
At a fixed state, the first $M$ rows of $\bm A_{\mathrm{ML}}$ are fixed, and the matrix is lower Hessenberg with nonzero superdiagonal entries.
Consequently, coefficient matching gives an affine, invertible relation between $\bm a$ and $\bm\beta$ \cite[Section~2.2.3]{Christlieb2025a}.
The two polynomial bases are related by an invertible triangular transformation.
Together, these relations give the forward map
\begin{equation*}
    \bm{\mathcal N}
    \longrightarrow
    \bm a
    \longrightarrow
    \bm\beta
    \longrightarrow
    \bm c
    \longrightarrow
    (\widetilde r_0,\ldots,\widetilde r_M).
\end{equation*}
Here $\widetilde r_j=\lambda_j-u$ are the roots of $P_{\bm a}$, determined up to permutation, and $\lambda_j$ are the characteristic speeds.

The neural closure uses this relationship in reverse.
The network predicts real shifted roots from $(\rho,\theta,f_3,\ldots,f_M)$, excluding $u$ to preserve Galilean invariance of the transport closure.
These roots define the monic polynomial
\begin{equation}
    P(y)=\prod_{j=0}^{M}(y-\widetilde r_j)
    =\sum_{k=0}^{M+1}c_k y^k,
    \qquad c_{M+1}=1.
    \label{eq:vieta-polynomial}
\end{equation}
Vieta's formulas give the coefficients $c_k$, and conversion to the temperature-scaled Hermite basis yields \cite[Section~2.2.4]{Christlieb2025a}
\begin{equation}
    \beta_i
    =\sum_{m=0}^{\lfloor(M+1-i)/2\rfloor}
      \frac{(i+2m)!}{2^m m!\,i!}\theta^m c_{i+2m},
    \qquad 0\leq i\leq M+1.
    \label{eq:monomial-to-hermite}
\end{equation}
Matching these Hermite coefficients with \eqref{eq:shifted-characteristic-polynomial} uniquely determines $\bm a$.
Finally, the gradient-closure coefficients are recovered from
\begin{equation}
    \mathcal N_i(\bm w)
    =\frac{a_i-(a_{\mathrm{Grad}})_i}{M+1},
    \qquad 0\leq i\leq M.
    \label{eq:last-row-to-gradient-coefficients}
\end{equation}
This reverse transformation is deterministic and is evaluated using the unstandardized physical state $\bm w$, including the temperature $\theta$.

The characteristic speeds are
\begin{equation}
    \lambda_j=u+\widetilde r_j,
    \qquad 0\leq j\leq M.
    \label{eq:physical-characteristic-speeds}
\end{equation}
Distinct real shifted roots give a real diagonalizable transport matrix and hence strict hyperbolicity.
Section~\ref{subsec:root-gap} describes how we enforce root separation.

\section{Training of the neural network and numerical implementation}
\label{sec3:training-and-numerics}

In this section, we describe the preparation of training and testing data and the training procedure for the neural closure.
To ensure eigenvalue separation, we introduce a prescribed minimum gap between the predicted characteristic roots.
Finally, we describe the common moment solver, including the path-conservative discretization, Poisson solve, and time integration.

\subsection{Data preparation}
We generate kinetic reference trajectories following the data-generation strategy of \cite{Han2019,Christlieb2025a}.
Spatial transport and electric acceleration are discretized using fifth-order finite-difference weighted essentially non-oscillatory (WENO5) schemes in physical and velocity space, respectively \cite{Jiang1996}.
The WENO stencils use periodic numerical boundary treatment in both variables.
Time integration uses the third-order ARS$(4,4,3)$ IMEX Runge--Kutta scheme \cite{Ascher1997}, with transport and electric acceleration treated explicitly and BGK relaxation implicitly.
At each explicit stage, the self-consistent electric field is computed from the current density using a Fourier solver for the periodic Poisson equation.

We use smooth initial data given by a weighted mixture of two local Maxwellians \cite{Han2019},
\begin{equation*}
    f(x,v,0)
    =\frac{\alpha_1\mathcal{M}(\rho_1,0,\theta_1)
    +\alpha_2\mathcal{M}(\rho_2,0,\theta_2)}
    {\alpha_1+\alpha_2+10^{-6}},
\end{equation*}
where $\alpha_1$ and $\alpha_2$ are sampled independently from uniform distribution $\mathcal{U}(0,1)$, and
\begin{equation*}
    \rho_i(x)=b_{i}+a_{i}
    \sin(2\pi k_{i}x+\phi_{i}),
    \qquad
    \theta_i(x)=\widetilde b_i+\widetilde a_i
    \sin(2\pi\widetilde k_i x+\widetilde\phi_i),
    \qquad i=1,2.
\end{equation*}
The amplitudes $a_i,\widetilde a_i$, constant offsets $b_i,\widetilde b_i$, and phases $\phi_i,\widetilde\phi_i$ are sampled uniformly from $[0.2,0.3]$, $[0.5,0.7]$, and $[0,2\pi]$, respectively.
The density and temperature parameter families are sampled independently.
For each family, the two wave numbers $k_i$ and $\widetilde{k}_i$ are sampled uniformly without replacement from $\{1,2,3,4\}$.
Each trajectory has a constant relaxation time sampled log-uniformly, $\log_{10}\tau\sim\mathcal{U}(-3,1)$.

\begin{table}[!htb]
    \centering
    \caption{Kinetic-data settings.}
    \label{tab:training-data}
    \begin{tabular}{ll}
        \toprule
        Parameter & Value \\
        \midrule
        Spatial grid & $N_x=256$, $x\in[0,1]$ \\
        Velocity grid & $N_v=512$, $v\in[-10,10]$ \\
        Time sampling & $T=0.3$, $\Delta t=2.5\times10^{-4}$, $\Delta t_{\rm out}=10^{-3}$ \\
        Training / test windows & $[0.051,0.15]$ / $[0.051,0.3]$ \\
        Stored moment variables & $15$ \\
        \bottomrule
    \end{tabular}
\end{table}

At each saved time, the macroscopic variables and Hermite coefficients are computed using composite midpoint quadrature in velocity, and their spatial derivatives are approximated using a fifth-order finite-difference WENO operator.
These quantities provide the retained states, their spatial gradients, and the target gradient $\partial_x f_{M+1}$ used in the closure loss.

Of the 1000 kinetic trajectories, 800 provide data for network training and validation, with their space-time samples randomly partitioned into $80\%$ for training and $20\%$ for validation.
The remaining $200$ form an evaluation set used to select the optimizer, number of retained moments, and minimum root gap, and to compare the selected neural closure with classical closures.

The discretization parameters and sampling windows are summarized in Table~\ref{tab:training-data}.

\subsection{Neural-network architecture and training}

\begin{figure}[!htb]
    \centering
    \includegraphics[width=0.75\textwidth]{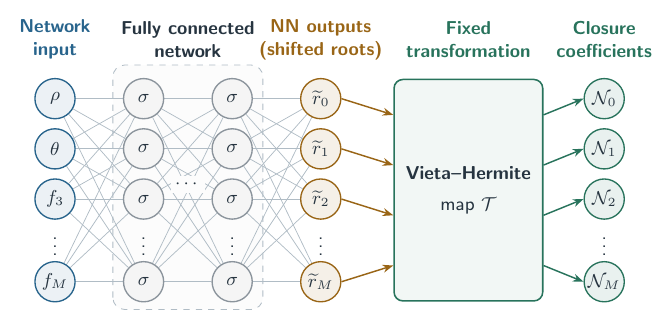}
    \caption{Neural network architecture.
    Here $\sigma$ denotes the ReLU activation in the hidden layers.}
    \label{fig:fcn-architecture-direct}
\end{figure}

Figure~\ref{fig:fcn-architecture-direct} illustrates the fully connected neural network (FCN) $\mathcal R_{\vartheta}:\mathbb R^M\to\mathbb R^{M+1}$, where $\vartheta$ denotes the trainable parameters.

The network takes the standardized features
\begin{equation}
    \bm z(\bm w)
    =\operatorname{standardize}(\rho,\theta,f_3,\ldots,f_M)
    \in\mathbb R^M.
    \label{eq:nn-features}
\end{equation}
Omitting the bulk velocity $u$ makes the shifted roots independent of $u$ and preserves the Galilean invariance of the transport closure.

It has nine affine layers with hidden width $128$ and ReLU activations in the hidden blocks.
The final affine layer directly predicts the $M+1$ real shifted roots $(\widetilde r_0,\ldots,\widetilde r_M)$, with no output activation.
The fixed Vieta--Hermite transformation then converts these roots into the gradient-closure coefficients $(\mathcal N_0,\ldots,\mathcal N_M)$, giving
\begin{equation}
    \bm{\mathcal N}(\bm w)
    =\mathcal{T}\left(
    \mathcal R_{\vartheta}\!\left(\operatorname{standardize}(\rho,\theta,f_3,\ldots,f_M)\right);
    \bm w\right).
    \label{eq:network-closure-map}
\end{equation}
Here $\mathcal T$ is the fixed, nontrainable transformation in \eqref{eq:vieta-polynomial}--\eqref{eq:last-row-to-gradient-coefficients}, evaluated using the unstandardized physical state $\bm w$.

We use this architecture for the optimizer comparison in Section~\ref{subsec:adamw-vs-soap}.
Because the predicted roots can coincide, we impose a minimum gap between adjacent roots in Section~\ref{subsec:root-gap}.

For a minibatch $\mathcal B$, training minimizes the mean-squared residual of the gradient closure,
\begin{equation}
    \mathcal{L}_{\mathcal B}(\vartheta)
    =\frac{1}{|\mathcal B|}\sum_{n\in\mathcal B}\left|
    \left(\partial_xf_{M+1}\right)^{(n)}
    -\sum_{i=0}^{M}\mathcal N_i(\bm w^{(n)};\vartheta)
    \left(\partial_xw_i\right)^{(n)}
    \right|^2,
    \label{eq:training-loss}
\end{equation}
where $w_i$ denotes the $i$th component of $\bm w=(\rho,u,\theta,f_3,\ldots,f_M)^{\mathsf T}$, indexed from $0$ to $M$, and the dependence of $\mathcal N_i$ on $\vartheta$ is written explicitly.
All derivatives in \eqref{eq:training-loss} are computed with the same fifth-order WENO operator used during preprocessing.

Following \cite{Christlieb2025a}, all networks use batch normalization in the hidden layers and a one-cycle learning-rate schedule.
Batch normalization uses minibatch statistics during training and running statistics during evaluation.
The learning rate is updated after each minibatch, first increasing to $10^{-3}$ and then decreasing to $5\times10^{-5}$.

\subsubsection{Optimizer comparison and model selection}
\label{subsec:adamw-vs-soap}
\label{subsec:model-selection}
\label{subsec:example1-periodic}
Christlieb et al.~\cite{Christlieb2025a} used AdamW~\cite{Loshchilov2019AdamW} to train their BGK closure on kinetic data and reported lower training errors than with Adam.
Vyas et al.~\cite{Vyas2025SOAP} proposed SOAP, which applies AdamW updates in the eigenbasis of Shampoo's matrix-valued preconditioners.
This incorporates information on gradient correlations into the adaptive updates.
To select an optimizer and moment count for the VPBGK closure, we compare AdamW and SOAP with $6$--$10$ retained moments in terms of closure-fitting errors, solution accuracy, and simulation completion.
At each moment count, the data, network architecture, loss, and learning-rate schedule are held fixed.

Both optimizers use minibatches of $4096$ and weight decay $10^{-2}$.
Each model is trained for $501$ epochs, and the epoch-$500$ checkpoint is used for evaluation.

\begin{table}[!htb]
    \centering
    \caption{Training and validation relative $L^2$ closure errors at epoch $500$.}
    \label{tab:adamw-soap-training-errors}
    \small
    \begin{tabular}{@{}ccccc@{}}
        \toprule
        & \multicolumn{2}{c}{AdamW} & \multicolumn{2}{c}{SOAP} \\
        \cmidrule(lr){2-3}\cmidrule(l){4-5}
        Retained moments & Training & Validation & Training & Validation \\
        \midrule
        $6$  & $1.858\times10^{-1}$ & $1.902\times10^{-1}$ & $1.760\times10^{-1}$ & $1.810\times10^{-1}$ \\
        $7$  & $1.164\times10^{-1}$ & $1.195\times10^{-1}$ & $1.138\times10^{-1}$ & $1.169\times10^{-1}$ \\
        $8$  & $7.587\times10^{-2}$ & $7.732\times10^{-2}$ & $7.520\times10^{-2}$ & $7.663\times10^{-2}$ \\
        $9$  & $1.022\times10^{-1}$ & $1.026\times10^{-1}$ & $1.065\times10^{-1}$ & $1.068\times10^{-1}$ \\
        $10$ & $2.331\times10^{-1}$ & $2.335\times10^{-1}$ & $2.321\times10^{-1}$ & $2.325\times10^{-1}$ \\
        \bottomrule
    \end{tabular}
\end{table}

Table~\ref{tab:adamw-soap-training-errors} reports the relative errors $\|\partial_x f_{M+1}-\sum_{i=0}^{M}\mathcal N_i\partial_x w_i\|_2/\|\partial_x f_{M+1}\|_2$ on the training and validation sets.
For both optimizers, these errors decrease from six to eight retained moments and increase at nine and ten.
The 8-moment SOAP model gives the lowest training and validation errors among the configurations considered, although its advantage over AdamW is small.

\begin{figure}[!htb]
    \centering
    \includegraphics[width=0.85\textwidth]{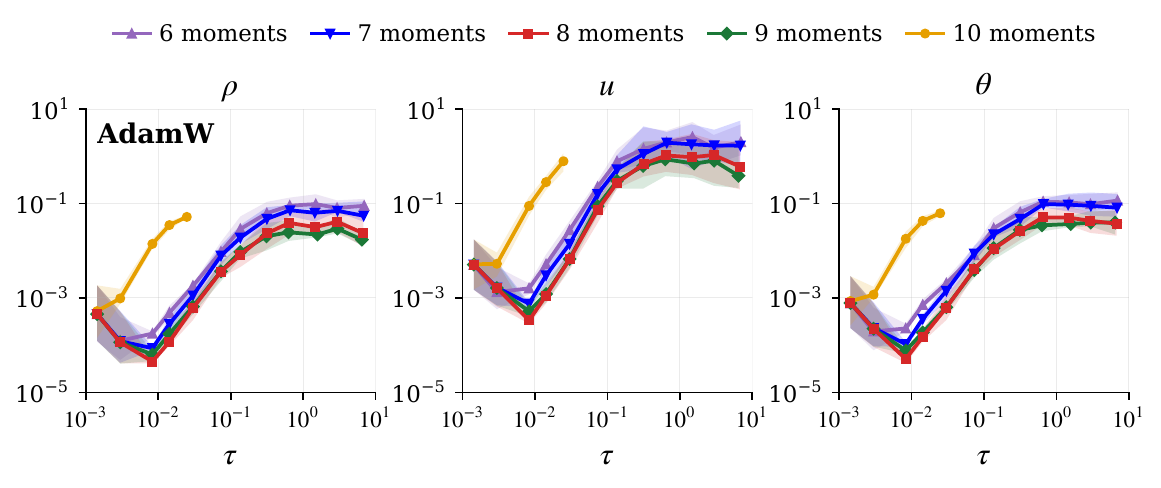}
    \par
    \includegraphics[width=0.85\textwidth]{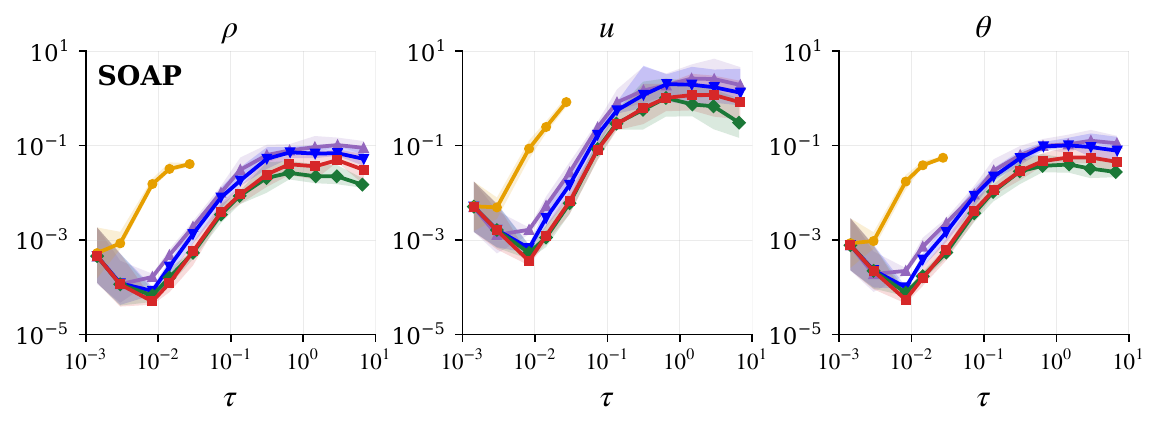}
    \caption{Relative $L^2$ errors in $\rho, u$, and $\theta$ at $t=0.3$ versus $\tau$ for AdamW (top) and SOAP (bottom), with $6--10$ retained moments.
    Curves and shading show geometric means and interquartile ranges in 12 common logarithmically spaced intervals of $\tau$, using each model's successful trajectories from the 200-case test set.
    Missing $10$-moment results at larger $\tau$ correspond to intervals with no successful simulations.
    See Figure~\ref{fig:example1-optimizer-failure-locations}.}
    \label{fig:example1-optimizer-errors}
\end{figure}

Figure~\ref{fig:example1-optimizer-errors} shows relatively small solution errors for $10^{-3}\leq\tau\leq10^{-2}$ and a general increase beyond this range.
Increasing the moment count does not uniformly improve accuracy; performance varies with $\tau$ and the state variable.

\begin{figure}[!htb]
    \centering
    \includegraphics[width=0.85\textwidth]{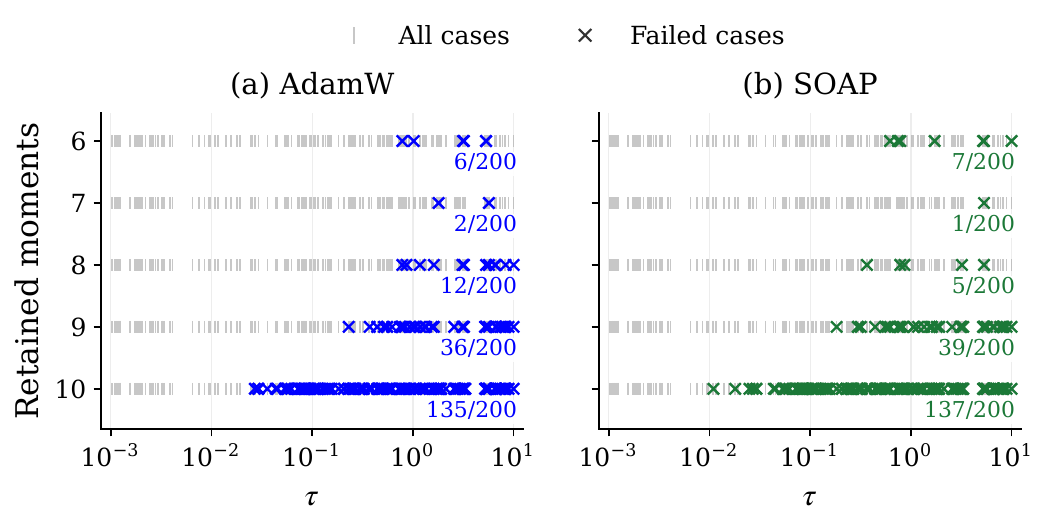}
    \caption{Locations of failed test simulations versus $\tau$ for (a) AdamW and (b) SOAP with $6$--$10$ retained moments.
    Gray marks show all 200 test cases, colored crosses mark failures.}
    \label{fig:example1-optimizer-failure-locations}
\end{figure}

Figure~\ref{fig:example1-optimizer-failure-locations} shows that failures are concentrated at larger relaxation times, with none observed for $\tau\leq10^{-2}$.
Failures become substantially more frequent at nine and ten retained moments.
With eight moments, SOAP completes $195$ of the $200$ test trajectories, compared with $188$ for AdamW.

\begin{figure}[!htb]
    \centering
    \includegraphics[ width=0.85\textwidth]{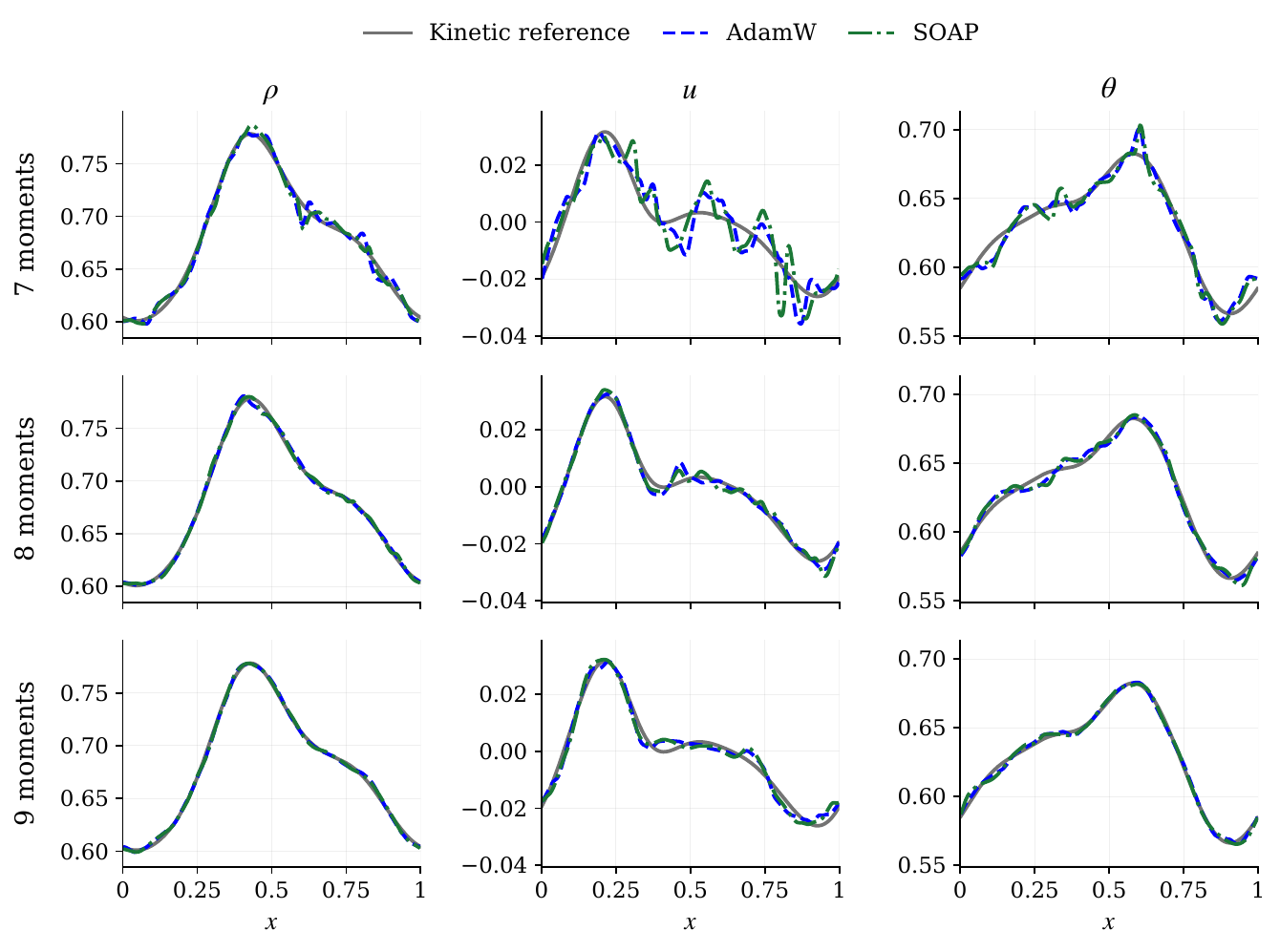}
    \caption{Moment profiles at $t=0.3$ for $\tau\approx0.1$ in the transition regime: kinetic reference, AdamW, and SOAP.
    Rows correspond to 7, 8, and 9 retained moments; columns show $\rho$, $u$, and $\theta$, with a common vertical scale within each column.}
    \label{fig:example1-moment-profiles}
\end{figure}

Figure~\ref{fig:example1-moment-profiles} compares moment profiles in the transition regime using the test trajectory with relaxation time closest to $\tau=0.1$.
For both optimizers, the seven-moment profiles exhibit spurious oscillations, particularly in velocity, whereas the eight- and nine-moment profiles agree more closely with the kinetic reference.
For this case, nine moments give smaller density and temperature errors, while eight moments give smaller velocity errors.

Considering closure-fitting accuracy, solution profiles, and trajectory completion, we select the 8-moment SOAP model for the subsequent root-separation study.

\subsubsection{Minimum characteristic-root separation}
\label{subsec:root-gap}

The 8-moment SOAP model in Section~\ref{subsec:adamw-vs-soap} still fails on five of the $200$ evaluation trajectories.
Because the original architecture does not enforce distinct roots, we impose a fixed minimum gap $\delta>0$ between the shifted roots supplied to the Vieta--Hermite transformation.

By the construction in Section~\ref{subsec:root-to-coefficient}, a positive minimum gap guarantees strict hyperbolicity of the full-gradient moment system when $\rho>0$ and $\theta>0$.
This structural guarantee alone does not ensure positivity preservation, stability, or accuracy of the numerical results.

\begin{samepage}
Let $(s_0,\ldots,s_M)=\mathcal R_{\vartheta}(\bm z)$ be the raw network outputs and let $s_{(0)}\leq\cdots\leq s_{(M)}$ denote their sorted values.
The roots supplied to the polynomial in \eqref{eq:vieta-polynomial} are
\begin{equation}
    \widetilde r_j
    =s_{(j)}+\delta\left(j-\frac{M}{2}\right),
    \qquad j=0,\ldots,M.
    \label{eq:gap-root-map}
\end{equation}
\end{samepage}
Consequently,
\begin{equation*}
    \lambda_{j+1}-\lambda_j
    =\widetilde r_{j+1}-\widetilde r_j
    =s_{(j+1)}-s_{(j)}+\delta
    \geq\delta,
    \qquad j=0,\ldots,M-1,
\end{equation*}
where $\lambda_j=u+\widetilde r_j$ as in \eqref{eq:physical-characteristic-speeds}.
Thus $\delta$ measures separation between shifted roots in the coordinate $\lambda-u$ and is not multiplied by $\sqrt{\theta}$.
The centered offsets sum to zero, preserving the mean of the raw roots while increasing their span by $M\delta$ for fixed raw outputs.

The layer in \eqref{eq:gap-root-map} is applied during both training and evaluation, and its output enters the deterministic root-to-coefficient map $\mathcal T$ in \eqref{eq:network-closure-map}.
Each positive-gap model is trained with its prescribed value of $\delta$ using the loss in \eqref{eq:training-loss}.
For $\delta=0$, the implementation bypasses sorting and offsets and recovers the original architecture.

\begin{figure}[!htb]
    \centering
    \includegraphics[width=0.85\textwidth]{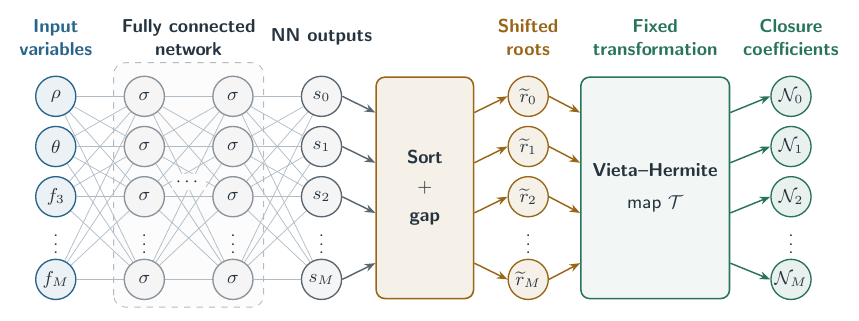}
    \caption{Neural network architecture with root separation.
    The raw network outputs are sorted and separated by a minimum gap $\delta$ to obtain shifted roots.}
    \label{fig:fcn-architecture}
\end{figure}

Figure~\ref{fig:fcn-architecture} shows the resulting architecture, with the sorting and gap layer inserted between the raw network outputs and the fixed transformation.

We compare 8-moment SOAP models trained with different prescribed gaps on the same $200$ evaluation trajectories.
All models use the numerical settings in Section~\ref{sec:num-method}, with characteristic speeds evaluated directly from the final predicted roots as $\lambda_j=u+\widetilde r_j$.
Table~\ref{tab:root-gap-selection} reports the resulting failure counts.

\begin{table}[!htb]
    \centering
    \caption{Failed simulations among the $200$ evaluation trajectories for 8-moment SOAP models with different minimum root gaps.}
    \label{tab:root-gap-selection}
    \small
    \begin{tabular}{@{}l*{10}{c}@{}}
        \toprule
        Gap $\delta$ & $0$ & $0.001$ & $0.01$ & $0.1$ & $0.15$ & $0.3$ & $0.5$ & $0.6$ & $0.7$ & $0.75$ \\
        \midrule
        Failures & $5$ & $9$ & $6$ & $9$ & $16$ & $11$ & $3$ & $0$ & $3$ & $16$ \\
        \bottomrule
    \end{tabular}
\end{table}

The failure count depends nonmonotonically on $\delta$: small gaps do not improve completion relative to the no-gap model, whereas $\delta=0.6$ gives zero failures and the best completion rate among the tested values for this ensemble.
The training and validation relative closure errors range approximately from $0.073$ to $0.089$ and from $0.075$ to $0.091$, respectively.
The gap-$0.6$ model minimizes neither error, showing that closure-fitting accuracy alone does not determine simulation completion.
Across the nine positive-gap models, $64$ simulations terminate when the closure matrix or characteristic speeds become nonfinite, and $9$ terminate when a stage temperature is nonfinite or at or below the admissibility threshold.
Thus, although the gap constraint ensures distinct real roots for admissible states in exact arithmetic, it neither bounds their spread nor guarantees admissibility or numerical stability.

\begin{figure}[!htb]
    \centering
    \includegraphics[width=0.85\textwidth]{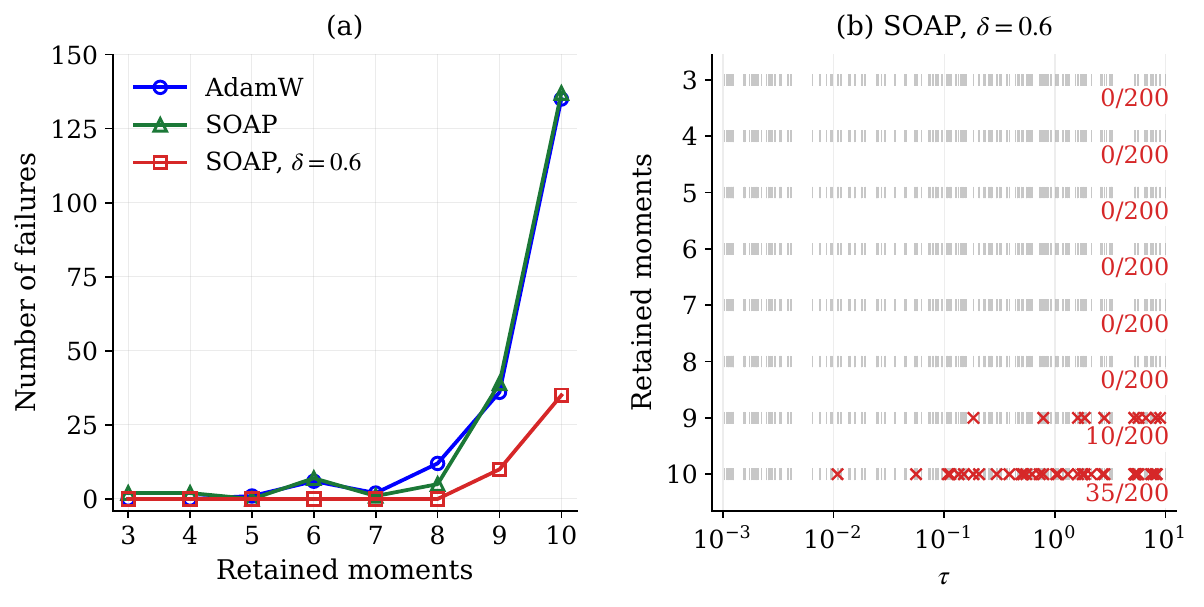}
    \caption{Failed simulations among the $200$ evaluation trajectories for epoch-$500$ models with $3$--$10$ retained moments: (a) failure counts for AdamW, SOAP, and SOAP with $\delta=0.6$; (b) failure locations for SOAP with $\delta=0.6$.
    In (b), gray ticks show all evaluation cases, red crosses mark failures, and annotations give failures out of $200$.}
    \label{fig:example1-optimizer-failures}
    \label{fig:example1-failure-locations}
\end{figure}

Figure~\ref{fig:example1-optimizer-failures} compares completion across $3$--$10$ retained moments.
The gap-$0.6$ models complete all $200$ trajectories through eight moments and reduce the nine- and ten-moment SOAP failure counts from $39$ to $10$ and from $137$ to $35$, respectively.
The remaining failures occur mainly at larger relaxation times, with some ten-moment failures also in the transition regime.
We therefore select the 8-moment SOAP model with $\delta=0.6$ for the subsequent neural-closure tests.
Appendix~\ref{app:root-gap-diagnostics} gives supplementary comparisons of eigenvalue distributions, transport-matrix condition numbers, and solution accuracy for paired simulations with and without root separation.

\subsection{Numerical implementation}
\label{sec:num-method}

We evolve the learned and classical moment systems using a path-conservative first-order centered (FORCE) finite-volume discretization \cite{KoellermeierCastro2021}, with WENO5 reconstruction \cite{Jiang1996} and SSPRK$(3,3)$ time integration \cite{Gottlieb1998}.
All closures share the spatial discretization, source treatment, and time-step policy, while their transport matrices and characteristic speeds differ.
Appendix~\ref{app:numerical-implementation} gives the variable transformations, closure matrices, quadrature, and complete stage operator.

At each Runge--Kutta stage, the reconstruction, closure matrices, Poisson field, and sources are recomputed.
The neural network is reevaluated at reconstructed nodal and interface-path states, and its final predicted shifted roots give the characteristic speeds $\lambda_j=u+\widetilde r_j$.
For periodic tests, the Poisson field is obtained by Fourier inversion of the current stage density; the slab boundary conditions are specified in Section~\ref{subsec:cross-regime-slab}.
The BGK and electric sources are treated explicitly without operator splitting.

For mesh width $\Delta x$ and closure $c$, the time step is
\begin{equation}
    \Delta t
    =\min\left\{
    0.4\frac{\Delta x}{a_c^n},
    0.25\tau_{\min},
    t_{\mathrm{out}}-t^n
    \right\},
    \label{eq:rollout-dt}
\end{equation}
where $a_c^n$ is $1.05$ times the largest absolute characteristic speed sampled at reconstructed interface-path quadrature points before the step, $\tau_{\min}=\min_j\tau(x_j)$, and $t_{\mathrm{out}}$ is the next output time.
The selected step remains fixed through all three stages.
A simulation terminates if a stage state or closure evaluation is nonfinite, or if a cell-state density or temperature is at or below $10^{-12}$.
For Grad, a relative imaginary part of the numerically computed spectrum exceeding $10^{-10}$ also causes termination.

\section{Numerical Tests}
\label{sec:numerical-tests}

Throughout this section, FCN denotes the eight-moment closure defined by the epoch-$500$ SOAP-trained network with minimum characteristic-root gap $\delta=0.6$.
The network parameters remain fixed in all tests, without retraining.
We first compare FCN with classical moment models using moment errors and Landau-damping field diagnostics.
We then assess its long-time robustness and its performance for two-beam relaxation with spatially varying collisionality and nonperiodic boundaries.
The comparison protocol is specified in each test, and the common numerical method is described in Section~\ref{sec:num-method}.

\subsection{Comparison with classical moment closures: Grad, HME, and QBME}
\label{subsec:classical-comparison}

We assess the accuracy and numerical robustness of FCN by comparing it with Grad, HME, and QBME against a kinetic reference.
All four models retain eight moments, corresponding to $M=7$ and $\bm w=(\rho,u,\theta,f_3,\ldots,f_7)^{\mathsf T}$ in \eqref{eq:primitive-hermite-state}.
The closure definitions are given in \eqref{eq:learned-gradient-closure}, \eqref{eq:three-closure-definitions}, and \eqref{eq:qbme-matrix-corrections}.

Both tests use the periodic domain $x\in[0,1]$ and the random two-Maxwellian initial-data family of Section~3.1, with a constant relaxation time $\tau$ for each trajectory.
The kinetic reference uses the scheme and resolution in Section~3.1 and Table~\ref{tab:training-data}.
All moment models use $N_x=256$ cells and the common FORCE--WENO5/SSPRK$(3,3)$ discretization, source treatment, and time-step controls of Section~\ref{sec:num-method}.
Each moment solution is initialized from the kinetic moments at $t=0.051$ and evolved independently to $t=0.3$, beyond the final training snapshot at $t=0.15$.

For a retained component $w$, we define the absolute and relative spatial $L^2$ errors at time $t$ by
\begin{equation}
    \begin{aligned}
        e_w(t)=\|w(t)-w^{\mathrm{ref}}(t)\|_{2,h},\quad
        \varepsilon_w(t)=\frac{e_w(t)}{\|w^{\mathrm{ref}}(t)\|_{2,h}}.
    \end{aligned}
    \label{eq:comparison-relative-error}
\end{equation}
Here $w^{\mathrm{ref}}(t)$ denotes the kinetic reference, and $\|\cdot\|_{2,h}$ is the discrete spatial $L^2$ norm.
In this subsection, the errors are evaluated at $t=0.3$.
Errors are reported only for simulations that reach the final time; termination criteria are specified in Section~\ref{sec:num-method}.

\subsubsection{Comparison on the 200-trajectory ensemble}
\label{subsec:classical-200-cases}

\begin{figure}[!htb]
    \centering
    \includegraphics[width=0.85\textwidth]{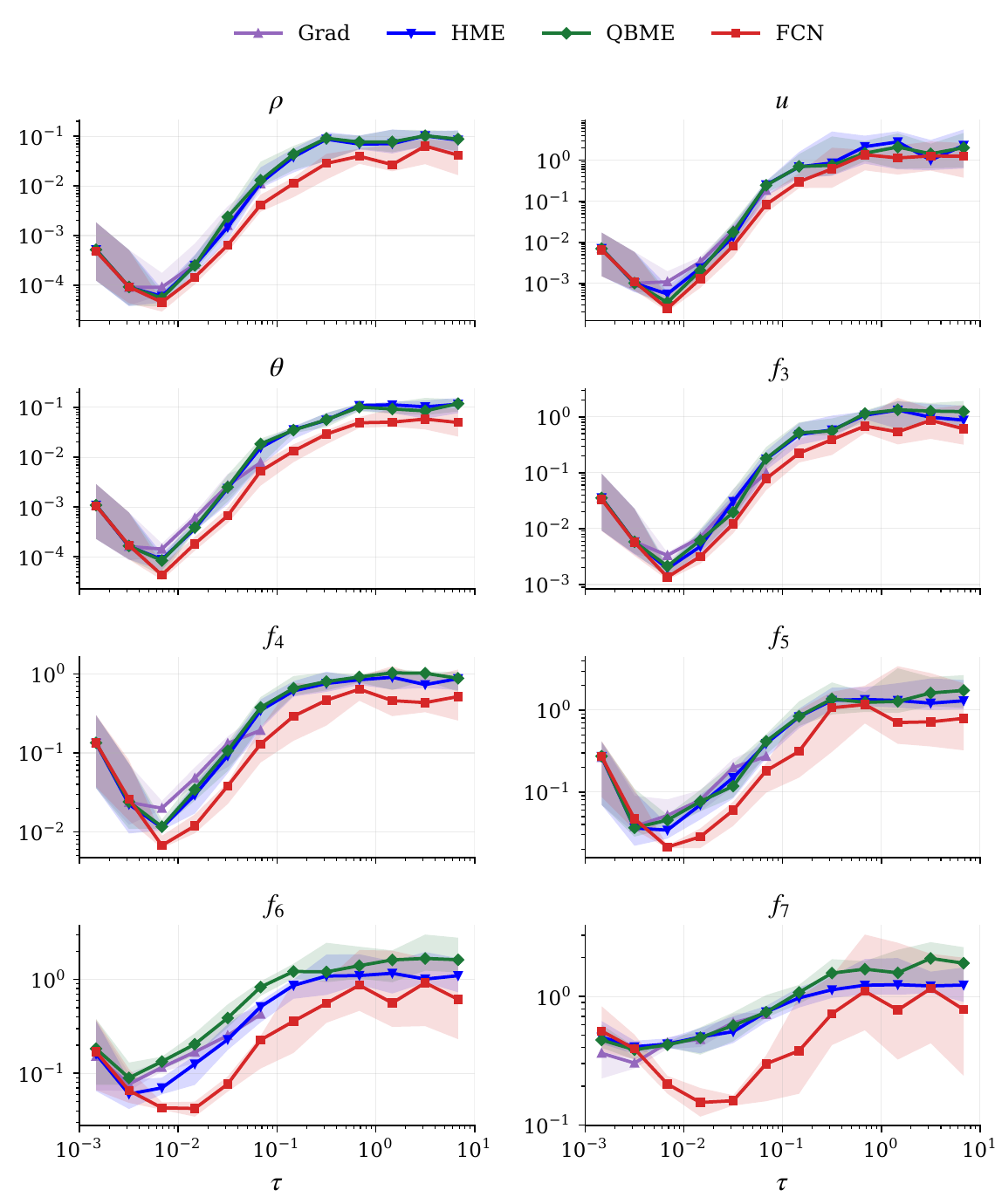}
    \caption{Relative $L^2$ errors at $t=0.3$ versus $\tau$ for all eight retained components.
    Solid curves with distinct markers show medians in $12$ common logarithmic bins; shading shows the $25$th--$75$th percentiles among each method's successful runs.
    Grad, HME, QBME and FCN are purple, blue, dark green and red, respectively.
    Empty bins are omitted, and the method-specific populations differ.}
    \label{fig:example1-closure-errors}
\end{figure}

We use the same $200$ trajectories, as in the neural-model comparisons of Section~\ref{subsec:model-selection}, with $\log_{10}\tau\sim\mathcal U(-3,1)$.
Figure~\ref{fig:example1-closure-errors} reports componentwise errors for all eight retained variables, summarized by medians and $25$th--$75$th percentiles in $12$ common logarithmic bins using each method's successful simulations.

In Figure~\ref{fig:example1-closure-errors}, we compare the relative $L^2$ errors of the four moment closures at $t=0.3$.
The results show that all four methods give comparable errors in $\rho$, $u$, and $\theta$ for small relaxation times, while the errors generally increase as $\tau$ increases beyond $10^{-2}$.
The Grad closure performs well in the strongly collisional regime, $\tau<10^{-2}$, but 131 of the 200 trajectories terminate after complex characteristic speeds are detected, and no simulations reach the final time for $\tau\geq0.1$.
The present one-dimensional Grad system is hyperbolic near local equilibrium but is not globally hyperbolic.
HME and QBME complete 200 and 199 trajectories, respectively, and give similar errors in the lower moments, while HME gives consistently smaller median errors in $f_4$ and $f_6$.
FCN completes all 200 trajectories and gives smaller median errors over most of the parameter range.
This improvement is particularly clear for $f_6$ and $f_7$ around $\tau=0.1$.
In the first interval above $\tau=0.1$, the median relative errors of HME and QBME approach order one, while those of FCN remain below $0.4$, giving approximately a threefold reduction.

\begin{figure}[!htb]
    \centering
    \includegraphics[width=0.85\textwidth]{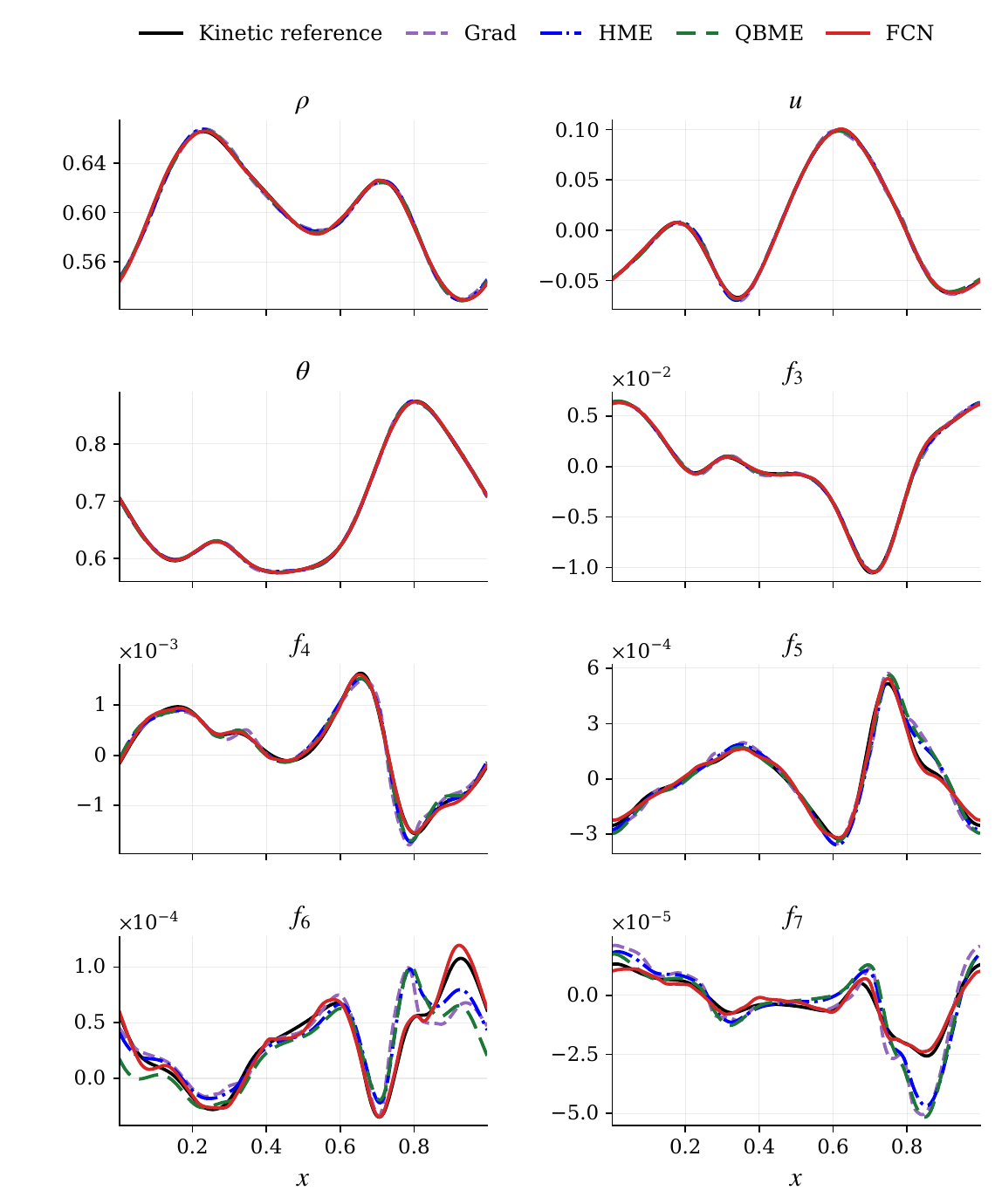}
    \caption{Eight retained moment profiles at $t=0.3$ for $\tau\approx0.036$.
    This case has the median FCN relative $L^2$ velocity error among the $43$ trajectories with $0.01\leq\tau\leq0.1$, selected independently of the other methods' completion.
    The kinetic reference is black; Grad, HME, QBME, and FCN are purple, blue, dark green, and red, respectively.}
    \label{fig:gap06-median-moments}
\end{figure}

\subsubsection{Comparison at fixed Knudsen numbers}
\label{subsec:classical-fixed-kn}
All methods perform well for the continuum regime, and here we are more interested in the performance of the proposed FCN in the transition regime, i.e., for relaxation time around $\tau=0.1$.
To do that, we generate $50$ new initial conditions and evolve each at $\tau\in\{0.1,0.2,0.3\}$, giving $150$ kinetic trajectories with initial conditions paired across the three Knudsen numbers.
At each Knudsen number, the representative case has the median FCN velocity error $\varepsilon_u$, ranked $26$th among the $50$ cases.
Selection is independent of the other methods' completion status and of the electric-field and heat-flux errors, and the selected initial-condition index may differ across Knudsen numbers.
We present the $\tau=0.1$ and $0.3$ results here.
The results for $\tau=0.2$ are similar to those for $\tau=0.1$ and are omitted for clarity.

For each selected case, we compare all eight retained profiles, the electric field reconstructed from each final density, and the heat flux $q=3f_3$ defined in \eqref{eq:heat-flux-f3}.

\begin{figure}[!htb]
    \centering
    \includegraphics[width=0.85\textwidth]{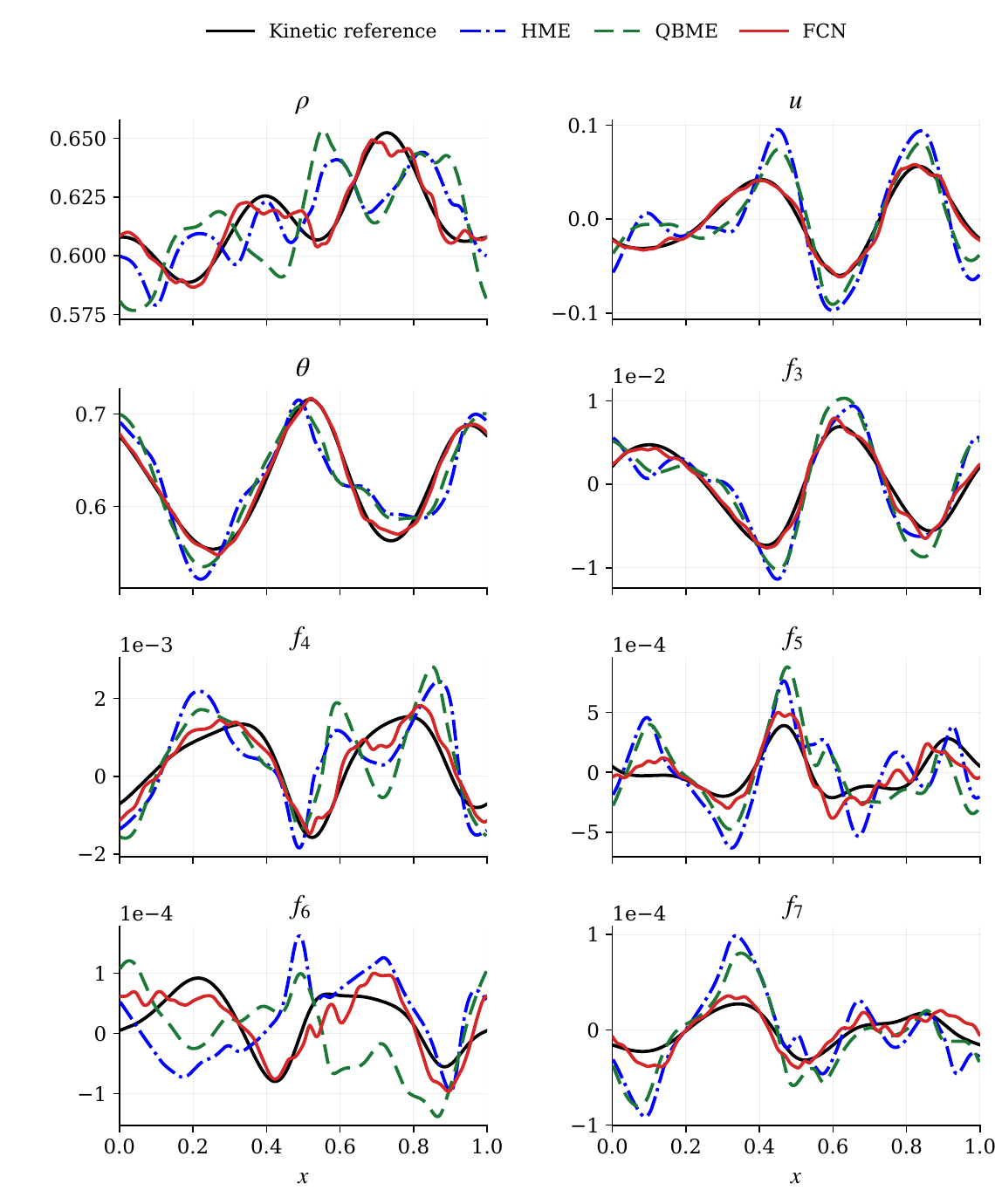}
    \caption{Comparison of the eight retained moments at $t=0.3$ for fixed $\tau = 0.1$.
    The case is selected by the upper-median FCN relative $L^2(u)$ error, rank $26$ of $50$.
    HME is blue dash-dotted, QBME dark green dashed and FCN red solid; the kinetic reference is black solid.}
    \label{fig:fixed-kn01-moments}
\end{figure}

\begin{figure}[!htb]
    \centering
    \includegraphics[ width=0.85\textwidth]{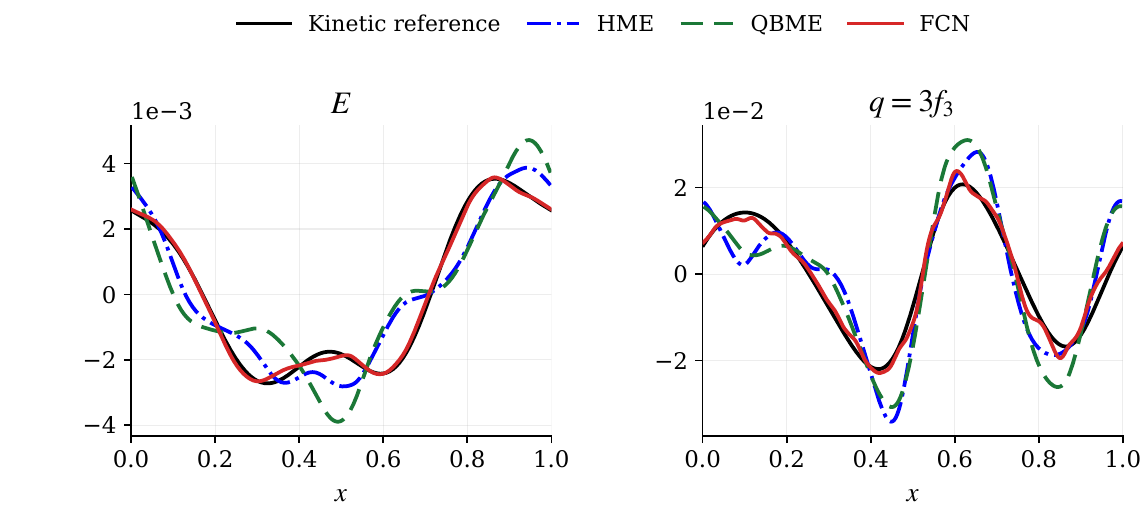}
    \caption{Electric field $E$ and heat flux $q=3f_3$ at $t=0.3$ for the same $\tau=0.1$ as Figure~\ref{fig:fixed-kn01-moments}.
    Selection uses only the FCN velocity error, independently of the two observables shown here.
    Colors and line styles match Figure~\ref{fig:fixed-kn01-moments}.}
    \label{fig:fixed-kn01-field-heat}
\end{figure}

For $\tau=0.1$, Figure~\ref{fig:fixed-kn01-moments} shows that FCN follows the kinetic reference more closely than HME and QBME across the retained moments.
In the density profile, HME and QBME introduce additional peaks and displace the main extrema, whereas FCN more accurately reproduces the broad spatial variations of the reference.
The two classical closures give similar profiles for $u$, $\theta$, and $f_3$, but tend to exaggerate the magnitudes of the main peaks and troughs of $u$ and $f_3$.
Their differences become more pronounced in the higher moments, particularly $f_6$, for which QBME predicts the wrong sign over parts of the domain.
FCN retains closer overall agreement in these components, although small-scale oscillations remain.
In Figure~\ref{fig:fixed-kn01-field-heat}, FCN better reproduces the electric field and the heat flux peaks, while HME and QBME displace or exaggerate.

\begin{figure}[!htb]
    \centering
    \includegraphics[ width=0.85\textwidth]{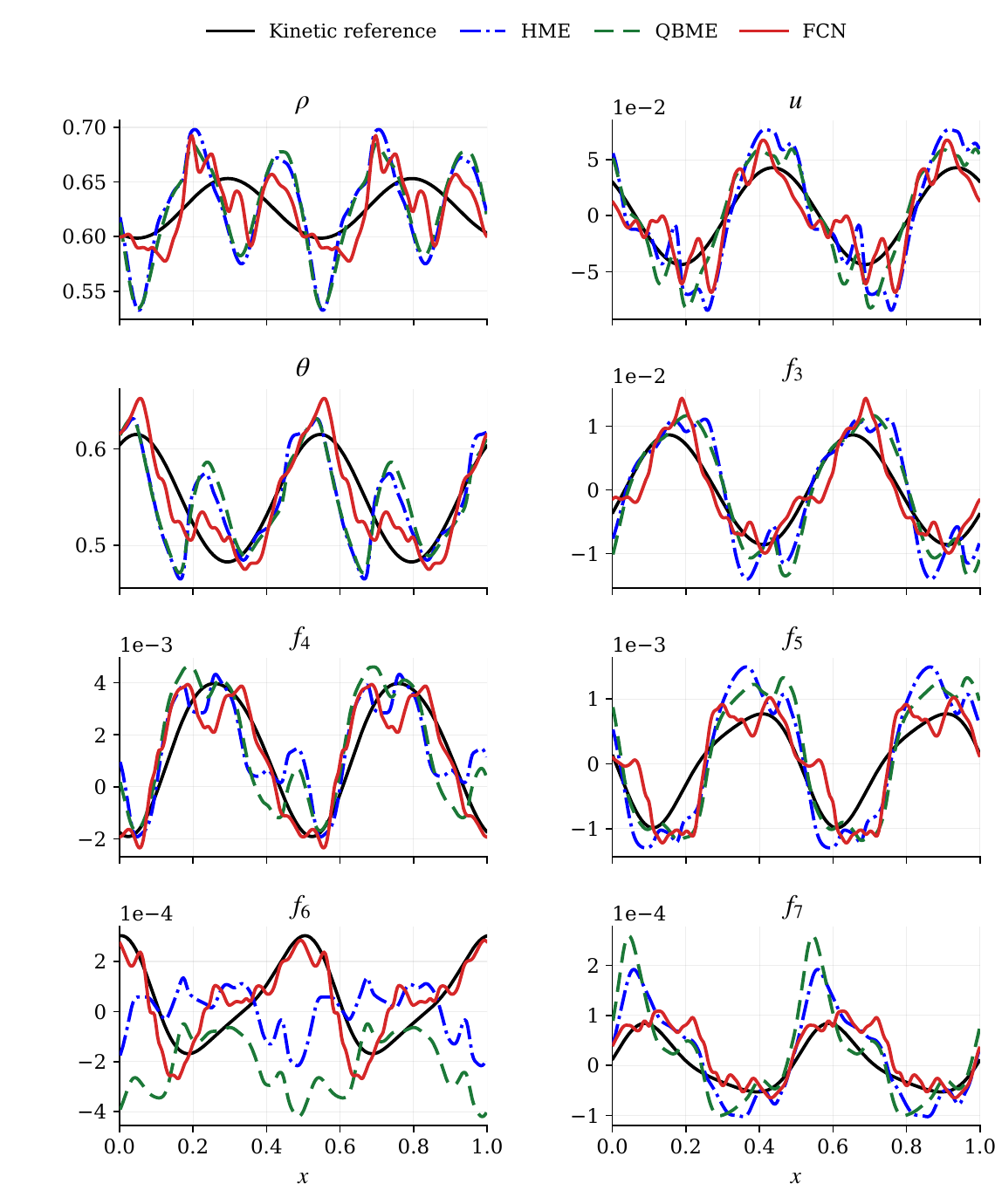}
    \caption{Eight retained moments at $t=0.3$ for $\tau=0.3$, selected independently within this $\tau$ ensemble by the upper-median FCN relative $L^2(u)$ error, rank $26$ of $50$.
    Colors and line styles match Figure~\ref{fig:fixed-kn01-moments}.}
    \label{fig:fixed-kn03-moments}
\end{figure}

\begin{figure}[!htb]
    \centering
    \includegraphics[ width=0.85\textwidth]{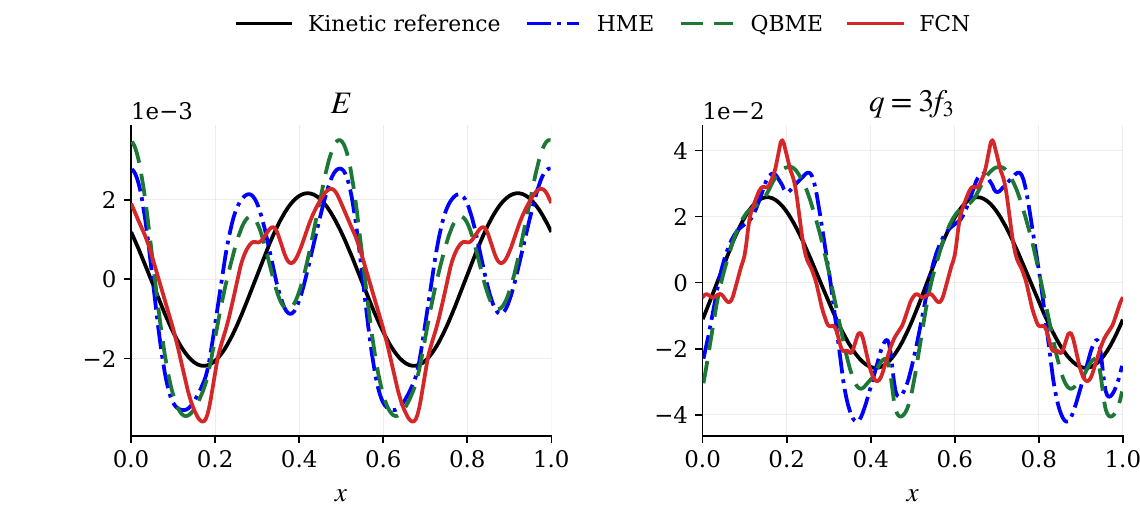}
    \caption{Electric field $E$ and heat flux $q=3f_3$ at $t=0.3$ for $\tau=0.3$, as selected for Figure~\ref{fig:fixed-kn03-moments}.
    Colors and line styles match Figure~\ref{fig:fixed-kn01-moments}.}
    \label{fig:fixed-kn03-field-heat}
\end{figure}

For the selected case in the less collisional regime $\tau=0.3$, Figure~\ref{fig:fixed-kn03-moments} shows noticeable additional spatial oscillations in all three moment closures relative to the smooth kinetic reference.
FCN nevertheless gives closer overall agreement in most components, with a particularly clear improvement in $f_6$, whose broad spatial variation is poorly reproduced by HME and QBME.
Figure~\ref{fig:fixed-kn03-field-heat} shows that FCN also better preserves the large-scale spatial pattern of the electric field.
HME and QBME introduce additional sign reversals within the positive lobes of the reference field, while FCN retains their positive sign, albeit with visible ripples and overly deep negative troughs.
The FCN heat-flux profile is also closer to the reference overall, but still overestimates the positive peaks and contains additional small-scale oscillations.

\par

\subsection{Landau damping and electrostatic observables}
\label{subsec:landau}

We use a Landau-damping benchmark to assess whether FCN reproduces the oscillation phase and amplitude decay of the electric field across different collisional regimes.
The problem is posed on the periodic domain $x\in[0,L]$, with $L=4\pi$, background density $\rho_{\mathrm b}=1$, and a mean-zero electric field.
The initial distribution and electric field are
\begin{equation}
    f(x,v,0)=\frac{1+\alpha\cos(kx)}{\sqrt{2\pi}}\exp\!\left(-\frac{v^2}{2}\right),
    \qquad E(x,0)=\frac{\alpha}{k}\sin(kx),
    \qquad \alpha=10^{-2},\quad k=\frac12.
    \label{eq:landau-data}
\end{equation}
The initial moments satisfy $u_0=0$, $\theta_0=1$, and $f_j(x,0)=0$ for $j\geq3$ in the moving Hermite basis.
We use constant relaxation times $\tau\in\{0.01,0.1,1,10\}$ and integrate over $0\leq t\leq20$, saving the solutions every $10^{-2}$ time units.

FCN, Grad, HME, and QBME retain eight moments and use the scheme of Section~\ref{sec:num-method} on $N_x=256$ cells, starting from the same exact initial cell averages.
The kinetic reference uses the scheme of Section~3.1 with the same spatial resolution, $N_v=512$ on $v\in[-10,10]$, and adaptive time stepping with CFL coefficient $0.8$.
For comparison, its partially conservative variables are cell-averaged before reconstructing the primitive moments and electric field.

\begin{figure}[!htb]
    \centering
    \includegraphics[ width=0.85\textwidth]{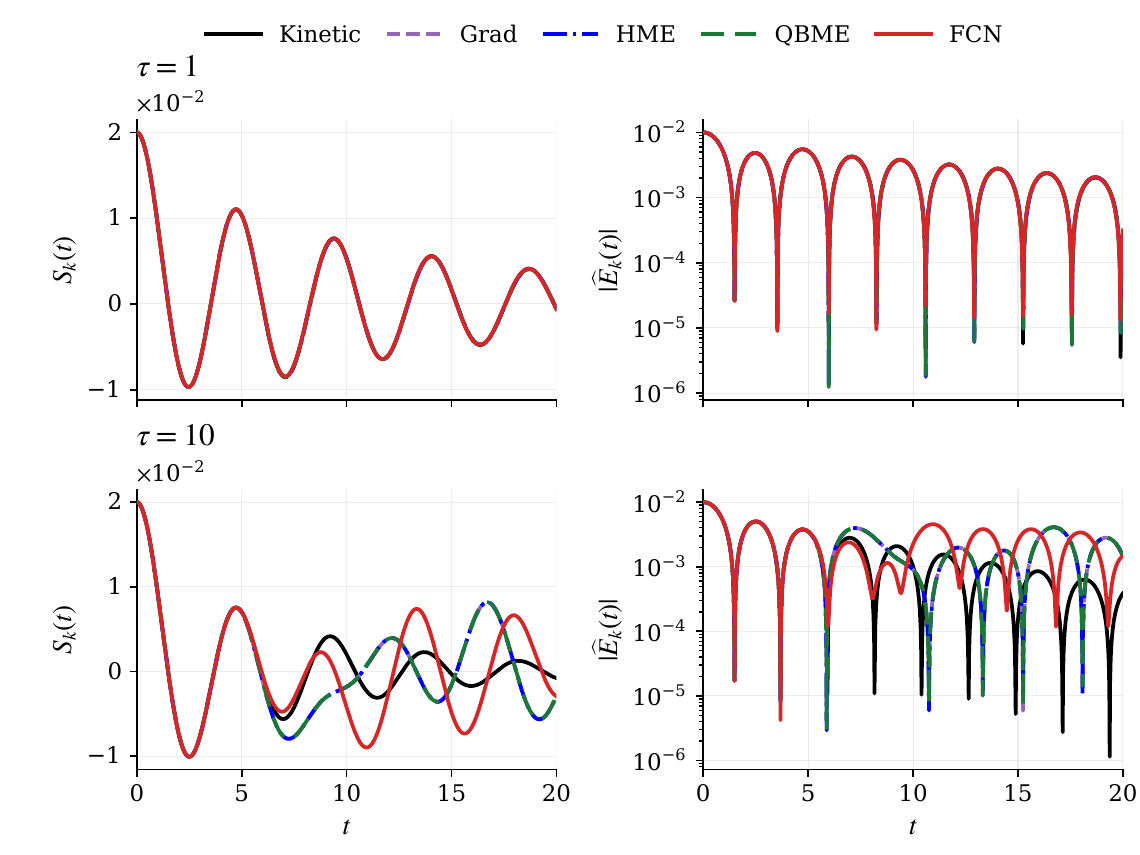}
    \caption{Landau-damping field histories for $\tau=1$ (top) and $\tau=10$ (bottom).
    Left: signed first-mode coefficient $S_k(t)$; right: magnitude $|\widehat E_k(t)|$ on a logarithmic scale.
    The curves compare FCN, Grad, HME, and QBME with the single-resolution kinetic reference.}
    \label{fig:landau-field-modes}
\end{figure}

Figure~\ref{fig:landau-field-modes} compares the signed first sine-mode coefficient $S_k(t)$ and Fourier-mode magnitude $|\widehat E_k(t)|$, where
\begin{equation}
    \widehat E_k(t)=\frac1L\int_0^L E(x,t)e^{-ikx}\,\mathrm dx,
    \qquad S_k(t)=-2\operatorname{Im}\widehat E_k(t).
    \label{eq:landau-field-diagnostics}
\end{equation}

For $\tau=0.01$ and $0.1$, all four closures closely reproduce the kinetic electric-field histories throughout the simulated interval; these nearly overlapping comparisons are omitted for brevity.
Figure~\ref{fig:landau-field-modes} shows that close agreement persists at $\tau=1$, with all methods capturing both the signed oscillations and the decay of the field-mode magnitude.
At $\tau=10$, the methods agree during the initial oscillations but develop substantial discrepancies after approximately $t=5$.
Grad, HME, and QBME give nearly identical responses, with distorted oscillation timing and excessive late-time amplitudes.
FCN more closely preserves the sequence of oscillations, but also exhibits appreciable phase and amplitude errors and fails to reproduce the sustained decay of the kinetic reference.

\par

\subsection{Long-time extrapolation and robustness}
\label{subsec:long-time-extrapolation}

This test assesses temporal extrapolation of FCN beyond the training interval $0.051\leq t\leq0.15$.
The ensemble consists of $200$ new realizations of the random two-Maxwellian initial-data family in Section~3.1 on the periodic domain $x\in[0,1]$, with constant relaxation times sampled as $\log_{10}\tau\sim\mathcal U(-3,1)$.

Each moment solution is initialized from the corresponding kinetic moments at $t=0$ and evolved up to $t=4$ using the scheme and termination criteria of Section~\ref{sec:num-method} with $N_x=256$ cells.
The kinetic reference uses the discretization of Section~3.1 with $N_x=256$, $N_v=512$, and time step $2.5\times10^{-4}$.
Diagnostics are recorded at intervals of $0.1$ time units.

\begin{figure}[!htb]
    \centering
    \includegraphics[width=0.75\textwidth]{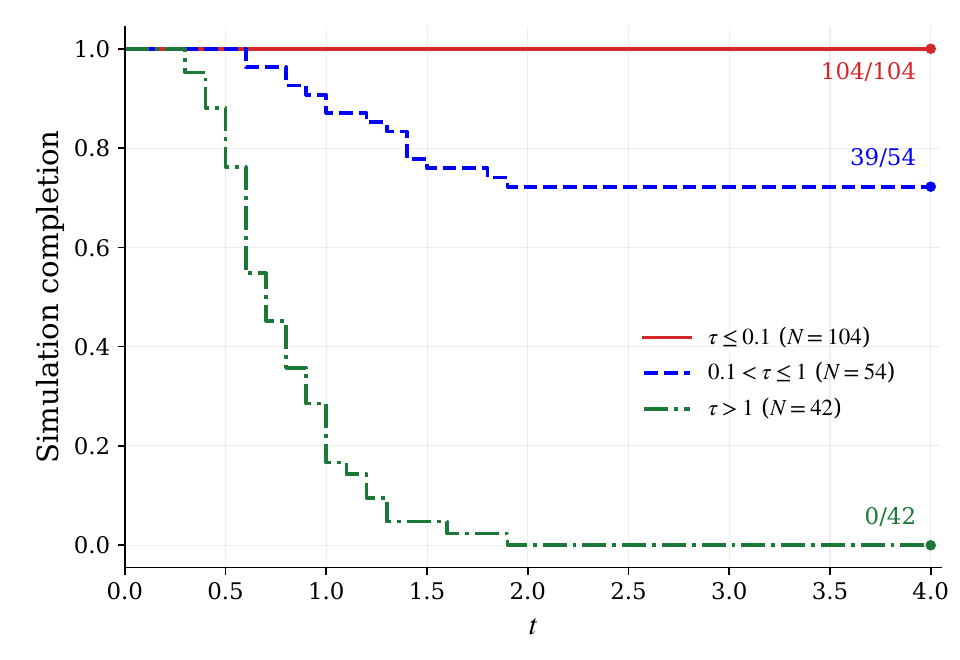}
    \caption{Simulation completion for FCN in the ranges $\tau\leq0.1$, $0.1<\tau\leq1$, and $\tau>1$.
    At each saved time $t$, the plotted fraction is the number of simulations reaching $t$ divided by the initial number in that range.}
    \label{fig:example1-long-time-completion}
\end{figure}

\begin{figure}[!htb]
    \centering
    \includegraphics[ width=0.85\textwidth]{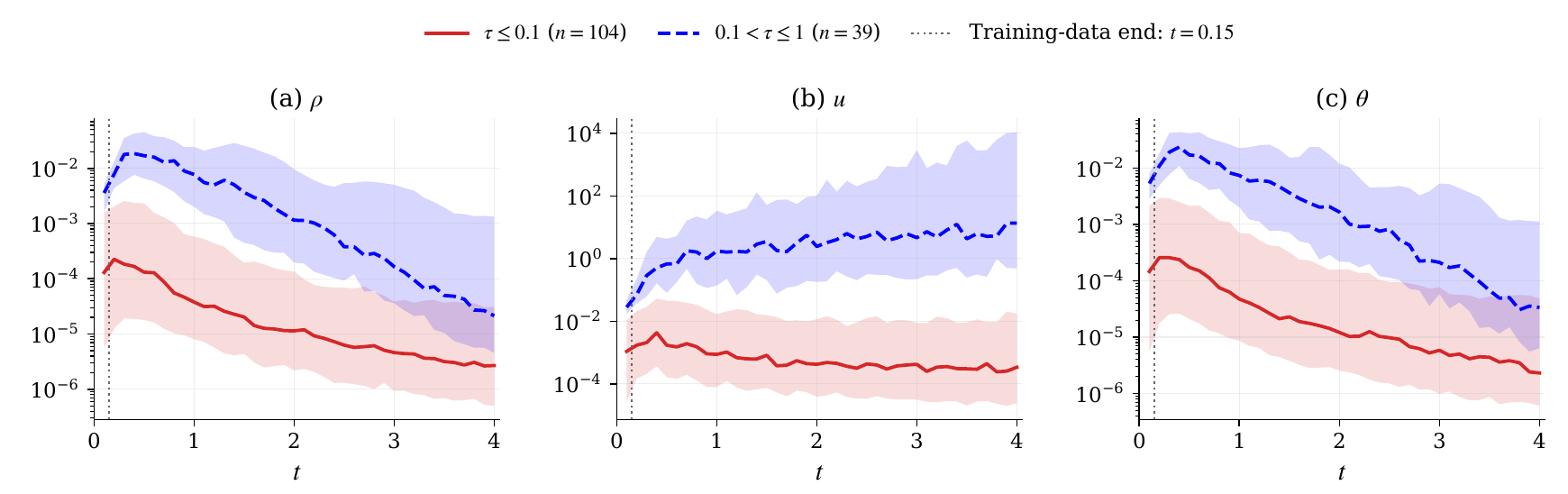}
    \caption{Long-time relative $L^2$ errors for $\rho$, $u$, and $\theta$ (left to right) using fixed cohorts of trajectories that reach $t=4$ in each indicated $\tau$ range.
    For $w\in\{\rho,u,\theta\}$, the relative error is $\|w-w^{\mathrm{ref}}\|_{2,h}/\|w^{\mathrm{ref}}\|_{2,h}$, where $\|w\|_{2,h}=(\Delta x\sum_j|w_j|^2)^{1/2}$ and $w^{\mathrm{ref}}$ is the kinetic reference.
    Curves give medians, shaded bands give the 10th--90th percentiles, and legend counts give the fixed cohort sizes.
    No error curve is shown for $\tau>1$ because all simulations in this range fail before $t=4$.
    The dotted lines mark the end of the training-data time window at $t=0.15$.}
    \label{fig:example1-long-time-errors}
\end{figure}

Figures~\ref{fig:example1-long-time-completion} and~\ref{fig:example1-long-time-errors} show the fraction of simulations reaching each saved time and the hydrodynamic error histories, respectively.
All $104$ simulations with $\tau\leq0.1$ reach $t=4$, as do $39$ of $54$ with $0.1<\tau\leq1$; the remaining $15$ fail before the final time.
All tested simulations with $\tau\leq0.3$ complete, whereas all $42$ simulations with $\tau>1$ fail before $t=2$.

Figure~\ref{fig:example1-long-time-errors} presents the relative errors in $\rho$, $u$, and $\theta$ for the long-time trajectories tests that reach $t=4$.
For $\tau\leq 0.1$, the median errors in all three variables remain small beyond the training interval (till $t=0.15$) and generally decrease at time progresses.
At $t=4$, the median density and temperature errors are of order $1\times10^{-6}$, while the median velocity error is approximately $3.4\times10^{-4}$.
For $0.1<\tau\leq 1$, the density and temperature errors are larger and exhibit greater variation across trajectories, but decrease at later times, reaching median errors around $2.1\times10^{-5}$, and $3.4\times10^{-5}$, respectively.
The relative velocity error instead increases, with a median of approximately $14$ at $t=4$ and a broad percentile band.
This increase is due to the decay of the reference velocity toward zero.
The absolute velocity error decreases at late times actually but remains large relative to the diminishing reference signal.
Thus, among the simulations with $0.1<\tau\leq1$ that reach $t=4$, the density and temperature approach the kinetic reference at late times, while the relative velocity error remains large.

\par

\subsection{Two-beam relaxation with spatially varying collisionality}
\phantomsection
\label{subsec:cross-regime-slab}

\begin{figure}[!htb]
    \centering
    \begin{tikzpicture}[x=10.2cm,y=1cm,font=\small]
        \fill[blue!65] (0,0) rectangle (0.25,2.0);
        \shade[left color=blue!65,right color=blue!10]
            (0.25,0) rectangle (0.5,2.0);
        \fill[blue!10] (0.5,0) rectangle (1,2.0);
        \draw[thick] (0,0) rectangle (1,2.0);
        \draw[densely dotted] (0.25,0) -- (0.25,2.0);
        \draw[densely dotted] (0.5,0) -- (0.5,2.0);

        \draw[very thick,dashed] (0,-0.12) -- (0,2.18);
        \draw[very thick,dashed] (1,-0.12) -- (1,2.18);
        \node[rotate=90,anchor=south] at (-0.018,1.0)
            {zero-gradient states};
        \node[rotate=90,anchor=north] at (1.018,1.0)
            {zero-gradient states};

        \node[text=white,font=\bfseries] at (0.125,1.62)
            {collisional};
        \node[font=\bfseries] at (0.375,1.62) {transition};
        \node[font=\bfseries] at (0.75,1.62) {weakly collisional};
        \node[text=white] at (0.125,0.90) {$\rho_0=0.8$};
        \node at (0.75,0.90) {$\rho_0=0.4$};
        \node[text=white] at (0.125,0.38)
            {$\tau=10^{-2}$};
        \node at (0.75,0.38) {$\tau=1$};

        \draw[->,thick] (0,-0.62) -- (1.08,-0.62) node[right] {$x$};
        \foreach \x/\lab in {0/0,0.25/{1/4},0.5/{1/2},1/1}
        {
            \draw (\x,-0.68) -- (\x,-0.56);
            \node[below] at (\x,-0.69) {$\lab$};
        }
        \node[above] at (0.5,2.15)
            {Electrostatic 1D1V model: $B=0$, $f=f(x,v,t)$};
    \end{tikzpicture}
    \caption{Simplified 1D1V electrostatic slab.
    The prescribed Knudsen number increases from $10^{-2}$ to $1$ and the initial density decreases from $0.8$ to $0.4$ across $1/4<x<1/2$.
    Shading distinguishes the collision regimes.
    Spatial boundaries use zero-gradient extrapolation for the kinetic and moment states, while the electric field satisfies $E(0,t)=E(1,t)=0$.}
    \label{fig:cross-regime-slab}
\end{figure}
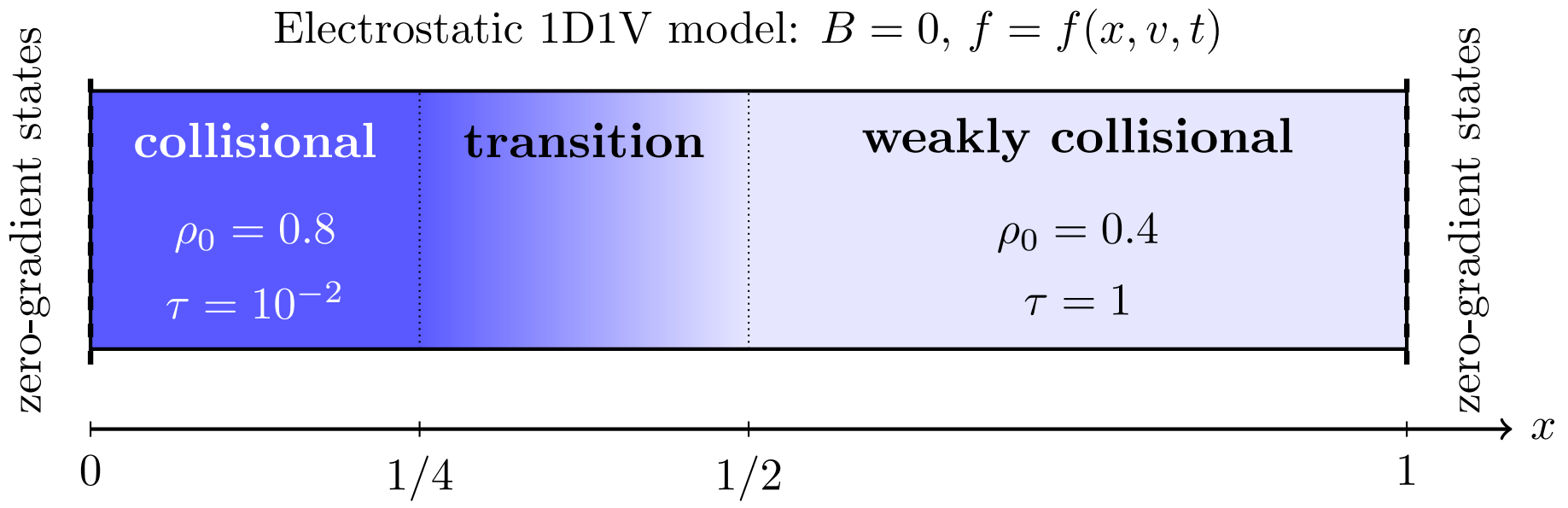

We consider the simplified Z-pinc motivated electrostatic slab \cite{shumlak2020z} in Figure~\ref{fig:cross-regime-slab} to test FCN with a two-beam initial distribution, spatially varying collisionality, and nonperiodic spatial boundaries.

The 1D1V model on $x\in[0,1]$ and $v\in\mathbb R$ is
\begin{equation}
    \partial_t f+v\partial_xf+E\partial_vf
    =\frac{\mathcal M[f]-f}{\tau(x)},
    \qquad
    \partial_xE=\rho-\rho_{\mathrm b}(t),
    \qquad
    \rho=\int_{\mathbb R}f\,\mathrm dv.
    \label{eq:cross-regime-vpbgk}
\end{equation}
We define the transition function
\begin{equation}
    \chi(x)=
    \begin{cases}
        0, & 0\leq x\leq1/4,\\
        3\xi^2-2\xi^3,\quad \xi=4x-1, & 1/4<x<1/2,\\
        1, & 1/2\leq x\leq1.
    \end{cases}
    \label{eq:cross-regime-transition}
\end{equation}
The prescribed relaxation time $\tau(x)$ and initial density are
\begin{equation}
    \tau(x)=10^{-2(1-\chi(x))},
    \qquad
    \rho_{\mathrm{init}}(x)=0.8-0.4\chi(x).
    \label{eq:cross-regime-profiles}
\end{equation}
These profiles connect a dense collisional region to a dilute, weakly collisional region across $1/4<x<1/2$.

Both solvers use zero-gradient ghost-cell extrapolation for the kinetic or moment state at the spatial boundaries.
At each time-integration stage, the electric field is reconstructed independently from each solver's density as
\begin{equation}
    E(x,t)=\int_0^x\bigl(\rho(y,t)-\rho_{\mathrm b}(t)\bigr)\,\mathrm dy,
    \qquad
    \rho_{\mathrm b}(t)=\int_0^1\rho(x,t)\,\mathrm dx.
    \label{eq:cross-regime-poisson-bc}
\end{equation}
This choice gives $E(0,t)=E(1,t)=0$, corresponding to homogeneous Neumann conditions for the electrostatic potential.

The initial distribution is an equal mixture of two Gaussian beams,
\begin{equation}
    f(x,v,0)=\frac{\rho_{\mathrm{init}}(x)}{2\sqrt{2\pi\theta_b}}
    \left[
        \exp\!\left(-\frac{(v-v_0)^2}{2\theta_b}\right)
        +\exp\!\left(-\frac{(v+v_0)^2}{2\theta_b}\right)
    \right],
    \qquad v_0=0.7,\quad\theta_b=0.31.
    \label{eq:slab-two-beam-data}
\end{equation}
The symmetric beams give $u_0=0$ and total temperature $\theta_0=\theta_b+v_0^2=0.8$.

Both solvers use $N_x=256$ spatial cells.
The kinetic reference follows the WENO5--IMEX$(4,4,3)$ discretization of Section~3.1, including the periodic velocity stencil, with $N_v=512$ points on $[-10,10]$ and $\Delta t=2.5\times10^{-4}$.
The moment solution is initialized from the kinetic moments at $t=0.051$ and evolved independently to $T=10$ using the discretization and time-step controls of Section~\ref{sec:num-method}, with the spatial boundary and field conditions specified above.
Both solutions are saved every $10^{-3}$ time units through $t=0.3$ and every $10^{-2}$ thereafter.

To quantify the profile differences over the full simulation, we evaluate the absolute and relative spatial $L^2$ errors defined in \eqref{eq:comparison-relative-error} at each saved time for all retained components $w\in\{\rho,u,\theta,f_3,\ldots,f_7\}$.

\begin{figure}[!htb]
    \centering
    \includegraphics[ width=0.85\textwidth]{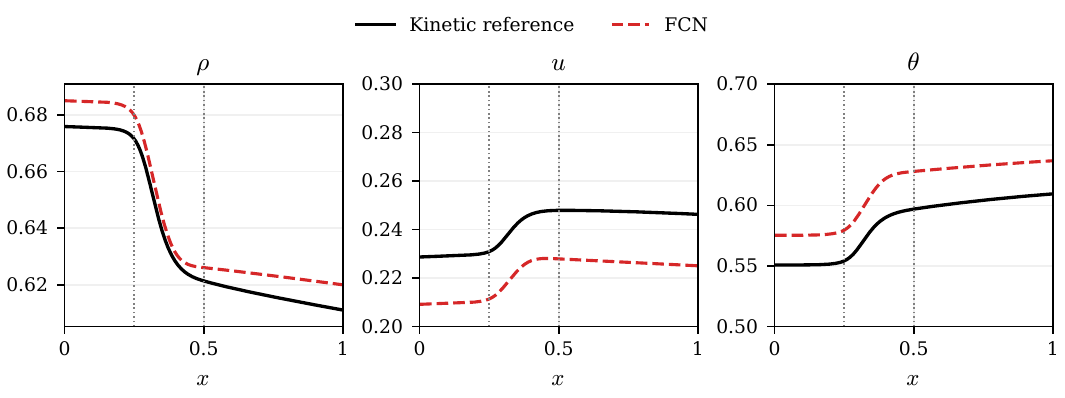}
    \caption{Density $\rho$, mean velocity $u$, and temperature $\theta$ at $t=10$ for the kinetic reference and FCN.
    Dotted lines mark the boundaries of the transition region $x\in[1/4,1/2]$.}
    \label{fig:slab-macro-t10}
\end{figure}

\begin{figure}[!htb]
    \centering
    \includegraphics[ width=0.85\textwidth]{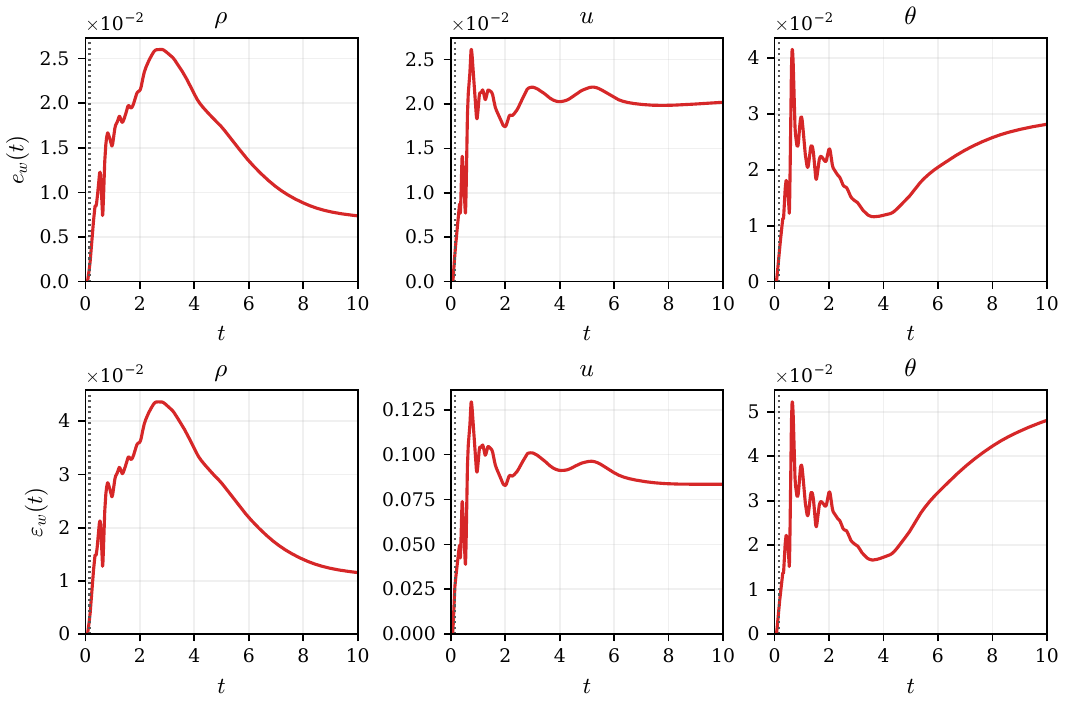}
    \caption{Absolute (top) and relative (bottom) spatial $L^2$ errors in $\rho$, $u$, and $\theta$ (left to right), comparing FCN with the kinetic reference.
    Errors are defined in \eqref{eq:comparison-relative-error}; dotted lines mark the training endpoint $t=0.15$.}
    \label{fig:slab-macro-errors}
\end{figure}

Figure~\ref{fig:slab-macro-t10} compares the density, mean velocity, and temperature at $t=10$.
The FCN reproduces the main spatial features of the kinetic reference, including the decrease in density and the increases in velocity and temperature across the transition region $1/4<x<1/2$.
Systematic offsets remain: the FCN overestimates the density and temperature and underestimates the mean velocity throughout the domain.
The final relative $L^2$ errors are approximately $1.16\times10^{-2}$, $8.35\times10^{-2}$, and $4.81\times10^{-2}$ for $\rho$, $u$, and $\theta$, respectively.

Figure~\ref{fig:slab-macro-errors} shows the evolution of the approximation errors during relaxation.
The absolute and relative $L^2$ errors, displayed in the top and bottom rows, respectively, exhibit similar temporal trends.
For all three variables, the errors initially grow from zero at $t=0.051$, when the test starts, but their subsequent evolution differs.
The density errors reach a maximum near $t=2.8$ and then decrease through $t=10$.
The mean velocity errors reach their maxima around $t=0.65$, then exhibit smaller variations, and remain approximately constant for $t\geq 8$.
The temperature errors reach a peak at approximately the same time as the velocity errors, then decrease with oscillations to a local minimum around $t=3.6$, and subsequently increase toward the final time.

\begin{figure}[!htb]
    \centering
    \includegraphics[ width=0.85\textwidth]{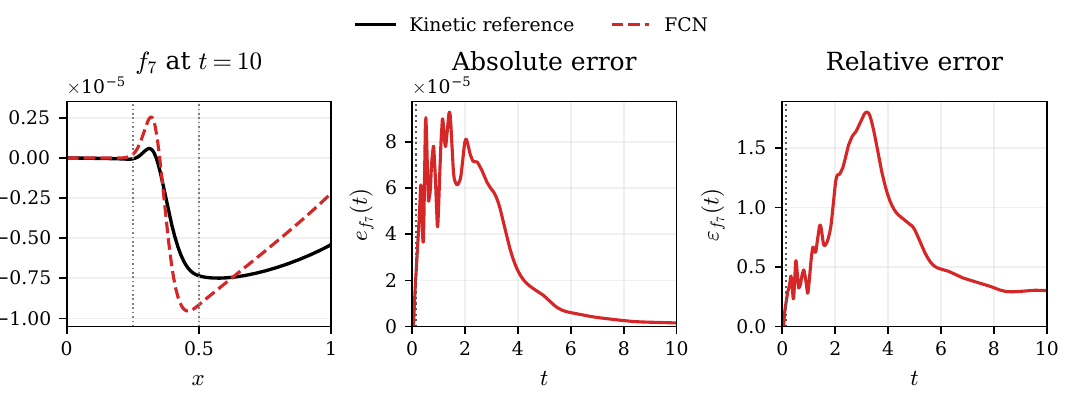}
    \caption{FCN versus the kinetic reference for the highest retained moment $f_7$: profiles at $t=10$ (left), absolute spatial $L^2$ error (middle), and relative error (right).
    Dotted lines mark the boundaries of the transition region $x\in[1/4,1/2]$ in the profile and the training endpoint $t=0.15$ in the error histories.}
    \label{fig:slab-f7-t10-errors}
\end{figure}

Figure~\ref{fig:slab-f7-t10-errors} compares the FCN prediction with the kinetic reference for the highest retained moment $f_7$ at $t=10$.
It also presents the corresponding absolute and relative $L^2$ error histories.
The FCN reproduces the sign change near the transition region, but predicts a larger positive peak and a deeper negative minimum than the kinetic reference.
The absolute $L^2$ error exhibits strong oscillations, reaching a maximum of approximately $9.29\times10^{-5}$ near $t=1.4$, before decreasing overall to $1.61\times10^{-6}$ at $t=10$.
The relative error peaks later, at approximately $1.8$ near $t=3.2$, and subsequently decreases overall to $0.303$ at the final time.
The absolute error decreases after the early transient, showing that the FCN prediction becomes closer to the kinetic reference.
However, the relative error continues to increase for some time because the reference $f_7$ becomes smaller faster than the error decreases.
At later times, both errors decrease overall, indicating improved agreement.
Nevertheless, the relative error remains around $0.3$ at $t=10$, so the remaining difference is still appreciable compared with the magnitude of the reference $f_7$.

\begin{figure}[!htb]
    \centering
    \includegraphics[ width=0.85\textwidth]{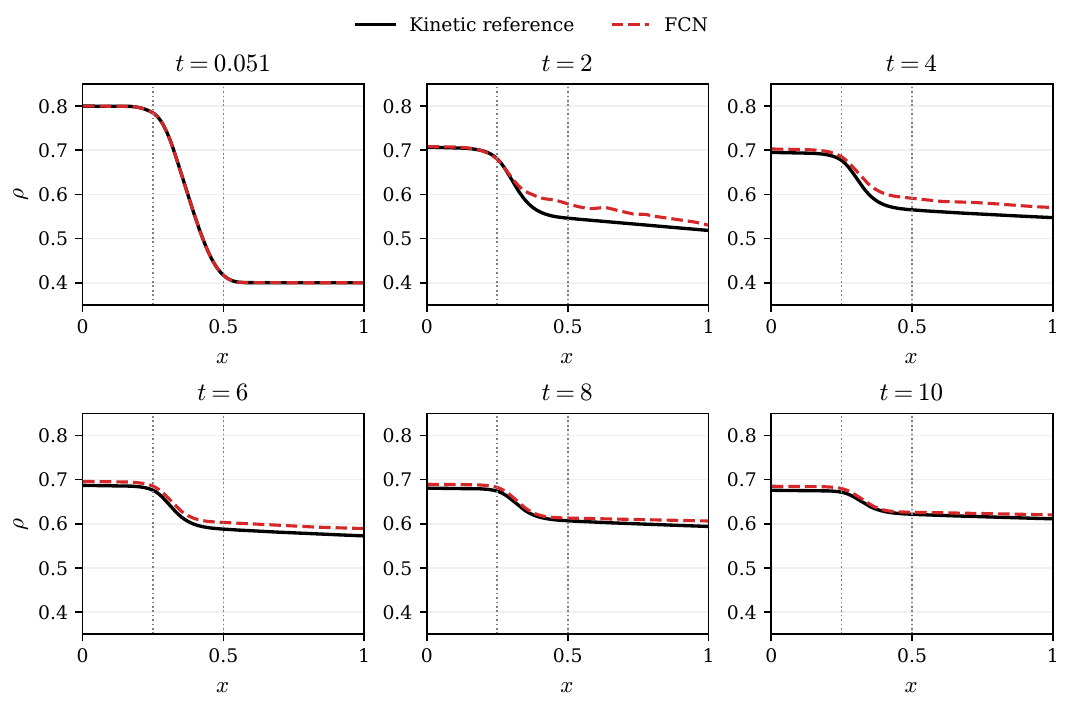}
    \caption{Density profiles at $t=0.051,2,4,6,8,10$, comparing the kinetic reference and FCN on a common vertical scale.
    Dotted lines mark the boundaries of the transition region $x\in[1/4,1/2]$.}
    \label{fig:slab-rho-evolution}
\end{figure}

Figure~\ref{fig:slab-rho-evolution} shows that the spatial variation in both density profiles decreases as time progresses.
The FCN captures this overall relaxation trend, although differences are visible at $t=2$ and $4$, particularly near the transition region.
Agreement improves at later times, with a slight overestimation of the density remaining at $t=10$.

\par

\section{Conclusion}\label{sec:conclusions}
We developed a hyperbolic learned moment closure for the VPBGK system by coupling a characteristic-root gradient closure to self-consistent electrostatic evolution.
For local, field-independent closure coefficients and a compatible periodic Poisson problem, eliminating the electric field leaves the principal transport matrix unchanged at a fixed admissible state.
The numerical results show that learning the closure can improve accuracy without increasing the number of evolved moments.
The selected eight-moment model gives smaller median errors than HME and QBME over most of the tested relaxation-time range, with approximately threefold reductions for the higher moments $f_6$ and $f_7$ in the interval $0.1\leq\tau\leq0.2$.
The root-separation study highlights the importance of assessing a learned closure through numerical evolution as well as closure-fitting errors.
The model with minimum gap $\delta=0.6$ completes all $200$ trajectories in the evaluation ensemble used for model selection.
With the trained network parameters fixed, we apply the closure without retraining to benchmark configurations absent from the training data, including Landau damping and a nonperiodic slab with spatially varying collisionality.
These tests extend far beyond the training endpoint $t=0.15$, reaching $t=4$ for temporal extrapolation, $t=10$ for the slab, and $t=20$ for Landau damping.
All tested long-time trajectories with $\tau\leq0.3$ reach $t=4$, and the nonperiodic slab test reaches $t=10$, with final relative errors of approximately $1.2\%$, $8.4\%$, and $4.8\%$ in density, mean velocity, and temperature, respectively.
These results support characteristic-root constraints as a practical approach to improving reduced VPBGK models, with predictive accuracy retained at low to moderate Knudsen numbers.
For $\mathrm{\tau}>0.1$, however, the remaining solution errors and failures in some long-time simulations show that further work is needed, even though the closure construction retains strict transport hyperbolicity at admissible states.

Future work will investigate how the prescribed root gap and the spread of characteristic speeds affect the conditioning of the characteristic decomposition and the regularity of the closure coefficients.
This analysis could guide additional constraints on the learned transport matrix and mechanisms for preserving positive density and temperature, with particular attention to weakly collisional regimes.
Further directions include extending the construction to higher-dimensional kinetic systems, where hyperbolicity must be considered in every spatial direction, and evaluating the resulting closures on multidimensional plasma benchmarks.

\section*{Acknowledgments}

The authors acknowledge the computational resources and support provided by the High Performance Computing Center (HPCC) of the Institute for Cyber-Enabled Research (ICER) at Michigan State University.
The authors also acknowledge support from ONR grant N00014-24-1-2242, NSF grant DMS-2618114, AFOSR grants FA9550-24-1-0254, DOE grant DE-SC0023164, and Ralph E. Powe Junior Faculty Enhancement Award from Oak Ridge Associated Universities (ORAU).

\appendix
\makeatletter
\renewcommand{\@seccntformat}[1]{\@ifundefined{#1@cntformat}{\csname the#1\endcsname\quad}{\csname #1@cntformat\endcsname}}
\newcommand{\section@cntformat}{Appendix~\thesection\quad}
\makeatother
\numberwithin{figure}{section}
\section{Spectral diagnostics for root separation}
\label{app:root-gap-diagnostics}

We supplement the root-gap study in Section~\ref{subsec:root-gap} by examining eigenvalue distributions and condition numbers of the transport matrix $\bm A_{\mathrm{ML}}$ in paired SOAP simulations with and without root separation.
The pairs include cases in which both models fail, only one completes, or both complete.
All plotted characteristic eigenvalues are evaluated directly as $\lambda_j=u+\widetilde r_j$ from the final predicted shifted roots.
In the cases examined, simulations that complete tend to show eigenvalues of the same locally sorted rank clustered together, with clearer separation between ranks when pooled over space.
Simulations that fail tend to show more overlap between these groups and larger matrix condition numbers.
These observed patterns do not constitute a stability guarantee.

\begin{figure}[!htb]
    \centering
    \includegraphics[width=0.85\textwidth]{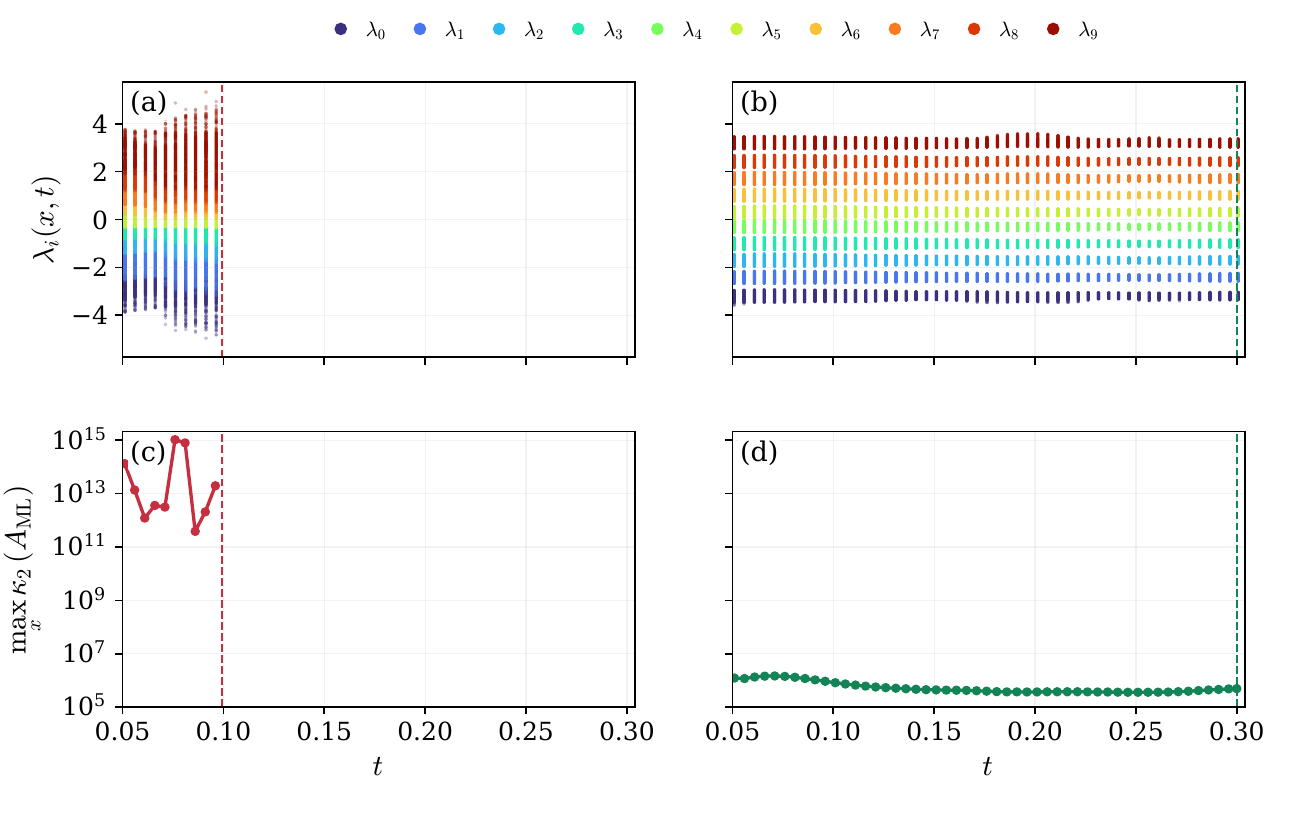}
    \caption{Characteristic eigenvalues (a,b) and spatial maximum matrix condition numbers $\kappa_{\max}(t)=\max_x\kappa_2(\bm A_{\mathrm{ML}}(x,t))$ (c,d) for ten-moment SOAP with $\tau=1$.
    The left and right columns correspond to the no-gap and gap-$0.6$ models, respectively.
    Colors identify locally sorted eigenvalue ranks over all $256$ spatial cells.
    Dashed lines mark the simulation stopping time.}
    \label{fig:case815-gap-analysis}
\end{figure}

Figure~\ref{fig:case815-gap-analysis} shows a ten-moment trajectory that terminates prematurely without a gap but completes with $\delta=0.6$.
The gap model exhibits clearer separation between eigenvalue groups and substantially smaller values of $\kappa_{\max}(t)$.
Because the plotted eigenvalues come from different spatial cells, their visual overlap does not by itself indicate a local eigenvalue collision.

Successful completion alone does not ensure accuracy.
In another 10-moment test, both the no-gap and gap-$0.6$ simulations reach $t=0.3$, but the no-gap solution exhibits substantial spatial oscillations.
At this time, its relative $L^2$ errors in $(\rho,u,\theta)$ are $(5.52\times10^{-2},4.144\times10^{-1},7.47\times10^{-2})$, compared with $(8.8\times10^{-4},2.65\times10^{-3},7.4\times10^{-4})$ for the gap model.
The improved results support root separation, while the remaining failures motivate further work on admissibility preservation and numerical robustness.

\section{Numerical implementation of the moment systems}
\label{app:numerical-implementation}

This appendix details the common numerical method summarized in Section~\ref{sec:num-method}.

\subsection{Moment variables and closure matrices}
\label{subsec:moment-variables-matrices}

For $M\geq3$, the solver advances cell averages of the partially conservative state
\begin{equation}
    \bm U=(\mu_0,\ldots,\mu_{M-1},f_M)^{\mathsf T},
    \qquad
    \mu_k=\int_{\mathbb R}v^k f\md v.
    \label{eq:partially-conservative-state}
\end{equation}
The change of variables between $\bm U$ and $\bm w$ is invertible for $\rho>0$ and $\theta>0$.
With $\bm J=\partial\bm U/\partial\bm w$, the transport matrix for closure $c\in\{\mathrm{Grad},\mathrm{HME},\mathrm{QBME},\mathrm{ML}\}$ becomes
\begin{equation}
    \widetilde{\bm A}^{(c)}(\bm U)
    =\bm J(\bm w)\bm A^{(c)}(\bm w)\bm J(\bm w)^{-1},
    \qquad \bm w=\bm w(\bm U),
    \label{eq:transformed-ml-matrix}
\end{equation}
and the transformed system is
\begin{equation}
    \partial_t\bm U+\widetilde{\bm A}^{(c)}(\bm U)\partial_x\bm U
    =\widetilde{\bm S}(\bm U,E),
    \label{eq:nonconservative-quasilinear}
\end{equation}
where the common transformed source is specified in Appendix~\ref{subsec:poisson_source_time}.

For Grad, HME \cite{Cai2012a}, and the neural closure, the coefficients in \eqref{eq:learned-gradient-closure} are
\begin{equation}
    \bm{\mathcal N}^{(\mathrm{Grad})}=\bm 0,\qquad
    \bm{\mathcal N}^{(\mathrm{HME})}
    =\left(0,-f_M,-\frac12f_{M-1},0,\ldots,0\right)^{\mathsf T},\qquad
    \bm{\mathcal N}^{(\mathrm{ML})}=\bm{\mathcal N}(\bm w).
    \label{eq:three-closure-definitions}
\end{equation}
Their primitive transport matrices satisfy
\begin{equation*}
    \bm A^{(c)}(\bm w)
    =\bm A_{\mathrm{Grad}}(\bm w)
    +(M+1)\bm e_{M+1}\bm{\mathcal N}^{(c)}(\bm w)^{\mathsf T},
    \qquad c\in\{\mathrm{Grad},\mathrm{HME},\mathrm{ML}\}.
\end{equation*}
For $M\geq4$, QBME instead uses $\bm A^{(\mathrm{QBME})}=\bm A_{\mathrm{Grad}}+\Delta\bm A$, with the three nonzero corrections \cite[Section~2.1.3]{KoellermeierCastro2021}
\begin{equation}
    \Delta A_{M,3}=-\frac{M(M+1)f_M}{2\theta},\qquad
    \Delta A_{M+1,3}=-\frac{M+1}{2}f_{M-1},\qquad
    \Delta A_{M+1,4}=\frac{3(M+1)f_M}{\rho\theta},
    \label{eq:qbme-matrix-corrections}
\end{equation}
where matrix entries use one-based indices in $(\rho,u,\theta,f_3,\ldots,f_M)$.
The transformation \eqref{eq:transformed-ml-matrix} is applied to the complete closure matrix, including both modified rows for QBME.
The characteristic speeds used to select the time step are
\begin{equation*}
    \lambda_\ell^{(c)}=
    \begin{cases}
        \lambda_\ell\!\left(\bm A_{\mathrm{Grad}}(\bm w)\right),
            & c=\mathrm{Grad},\\
        u+\sqrt{\theta}\,\zeta_\ell,
            & c\in\{\mathrm{HME},\mathrm{QBME}\},\\
        u+\widetilde r_\ell(\bm w),
            & c=\mathrm{ML},
    \end{cases}
    \qquad 0\leq\ell\leq M,
\end{equation*}
where $\zeta_\ell$ are the zeros of $\He_{M+1}$ and $\widetilde r_\ell$ are the final shifted roots, including the separation layer \eqref{eq:gap-root-map} when used.
For every neural model, the characteristic speeds are evaluated directly from these predicted shifted roots in all reported simulations.
For Grad, the eigenvalues are computed numerically from $\bm A_{\mathrm{Grad}}$, and a simulation is terminated if the relative imaginary part of its spectrum exceeds $10^{-10}$.

\subsection{Path-conservative spatial discretization}
\label{subsec:path_conservative_discretization}

The first $M$ transport equations in $\bm U$ are conservative for Grad, HME, and the neural closure; QBME modifies the two highest equations, leaving the first $M-1$ unchanged.
For all closures, we prescribe straight-line paths in $\bm U$ across interfaces.
Writing $\Delta\bm U=\bm U^+-\bm U^-$, the path, averaged matrix, and path jump are
\begin{equation}
    \begin{aligned}
        \bm\Phi(s;\bm U^-,\bm U^+)
        &=\bm U^-+s\Delta\bm U,\qquad 0\leq s\leq1,\\
        \widetilde{\bm A}^{(c)}_{\Phi}(\bm U^-,\bm U^+)
        &=\int_0^1\widetilde{\bm A}^{(c)}(\bm\Phi(s))\md s,\\
        \mathcal J^{(c)}(\bm U^-,\bm U^+)
        &=\widetilde{\bm A}^{(c)}_{\Phi}(\bm U^-,\bm U^+)\Delta\bm U.
    \end{aligned}
    \label{eq:path-jump-definition}
\end{equation}
At each path quadrature point, we recover $\bm w$ and evaluate the complete closure matrix; the electric field enters only through the source.

On a uniform mesh with cell centers $x_j$ and width $\Delta x$, componentwise WENO5 reconstruction from the cell averages $\overline{\bm U}_j$ supplies states $\bm U_{j,m}$ at $x_j+\Delta x\,\xi_m$.
Here $(\xi_m,\omega_m)_{m=1}^4$ are the nodes and weights of four-point Gauss--Lobatto quadrature on $[-1/2,1/2]$.
If primitive recovery at a reconstructed node gives a nonfinite density or temperature, or either is at or below $10^{-10}$, all four nodal states in that cell are replaced by its cell average.
The cubic interpolant $\bm P_j(x)$ through the resulting nodal states supplies the cell-interior derivative.

At each interface, the reconstructed traces define $\Delta\bm U_{j+1/2}$ and the path $\bm\Phi_{j+1/2}$ in \eqref{eq:path-jump-definition}.
Using three-point Gauss--Legendre nodes and weights $(s_r^{\mathrm{GL}},\omega_r^{\mathrm{GL}})$ on $[0,1]$, we approximate the averaged matrix by

\begin{equation*}
    \widetilde{\bm A}^{(c)}_{\Phi,j+1/2}
    :=\sum_{r=1}^{3}\omega_r^{\mathrm{GL}}
    \widetilde{\bm A}^{(c)}\!\left(\bm\Phi_{j+1/2}(s_r^{\mathrm{GL}})\right).
\end{equation*}
Following \cite{KoellermeierCastro2021}, we define the path-conservative FORCE fluctuations by
\begin{equation}
    \mathcal{D}_{j+1/2}^{\pm}
    =\frac12\left\{
    \widetilde{\bm A}^{(c)}_{\Phi,j+1/2}\Delta\bm U_{j+1/2}
    \mathbin{\pm}\frac12\left[
    \frac{\Delta x}{\Delta t}\bm I
    +\frac{\Delta t}{\Delta x}
    \left(\widetilde{\bm A}^{(c)}_{\Phi,j+1/2}\right)^2
    \right]\Delta\bm U_{j+1/2}\right\}.
    \label{eq:force-fluctuations}
\end{equation}

The cell-interior contribution and finite-volume stage operator are
\begin{equation}
    \begin{aligned}
        \mathcal I_j^{(c)}
        &=\Delta x\sum_{m=1}^{4}\omega_m
        \widetilde{\bm A}^{(c)}(\bm U_{j,m})
        \partial_x\bm P_j(x_j+\Delta x\,\xi_m),\\
        \mathcal L_{h,j}^{(c)}(\overline{\bm U};\Delta t)
        &=-\frac{1}{\Delta x}\left(
        \mathcal{D}_{j-1/2}^{+}+\mathcal{D}_{j+1/2}^{-}+\mathcal I_j^{(c)}
        \right)+\widetilde{\bm S}(\overline{\bm U}_j,E_j).
    \end{aligned}
    \label{eq:high_order_path_conservative_scheme}
\end{equation}

The source is evaluated at the cell-average state with the local relaxation time $\tau(x_j)$.
Periodic stencils are used in the periodic tests.

\subsection{\texorpdfstring{Poisson solve and time integration}{Poisson solve and time integration}}
\label{subsec:poisson_source_time}

At each SSPRK$(3,3)$ stage, we rebuild the reconstruction, closure matrices, Poisson field, and sources; for the neural closure, the network is reevaluated at the reconstructed nodal and path states.
For periodic tests, we solve \eqref{eq:vpbgk-poisson} by Fourier inversion using the current stage density and the fixed background $\rho_0$ defined in Section~\ref{sec2:model}.
The discrete mean charge is removed before inversion, and the zero and Nyquist modes of the electric field are set to zero.
The BGK and electric sources from \eqref{eq:vpbgk-sources} enter the stage residual explicitly, without operator splitting, through

\begin{equation*}
    \widetilde{\bm S}(\bm U,E)
    =\bm J(\bm w)\left[
    \bm Q_{\mathrm{BGK}}(\bm w;\tau)+\bm Q_E(E)
    \right],
    \qquad \bm w=\bm w(\bm U).
\end{equation*}
The slab tests retain the same interior discretization, with their Poisson boundary conditions and relaxation-time profiles specified in Section~\ref{subsec:cross-regime-slab}.

Before each step, the characteristic speeds are sampled at the interface-path quadrature points reconstructed from $\overline{\bm U}^{n}$.
The speed bound in \eqref{eq:rollout-dt} is
\begin{equation}
    a_c^n
    =1.05\max_{j,r,\ell}
    \left|\lambda_\ell^{(c)}\!\left(
    \bm w\!\left(\bm\Phi_{j+1/2}(s_r^{\mathrm{GL}})\right)
    \right)\right|.
    \label{eq:characteristic-speed-bound}
\end{equation}
The common time-step restrictions and termination criteria are specified in Section~\ref{sec:num-method}.

\bibliographystyle{abbrv}
\bibliography{references}

\end{document}